%% file: colm2024_conference.tex
\documentclass{article} 
\usepackage{colm2024_conference}

\input{math_commands.tex}

\usepackage{hyperref}
\usepackage{url}

\usepackage{amsmath}
\usepackage{graphicx}
\usepackage{color}
\usepackage{xcolor}
\usepackage{siunitx}
\usepackage{amsfonts}
\usepackage{latexsym}
\usepackage{comment}
\usepackage{booktabs}
\usepackage{microtype}
\usepackage{xspace}
\usepackage{booktabs}
\usepackage{multirow}
\usepackage{dcolumn}
\usepackage{tikz}
\usepackage{bbding}
\usepackage{hhline}
\usepackage{mdframed}
\usepackage{enumitem}
\usepackage{xcolor}
\usepackage{hyperref}
\usepackage{url}
\newcolumntype{d}[1]{D{.}{.}{#1}}
\usepackage{makecell}
\usepackage{dirtytalk}
\usepackage{colortbl}
\usepackage{wrapfig}
\usepackage{caption}
\usepackage{bm}
\usepackage{amssymb}
\usepackage{algorithm}
\usepackage[noend]{algpseudocode}
\definecolor{darkblue}{rgb}{0, 0, 0.5}
\hypersetup{colorlinks=true, citecolor=darkblue, linkcolor=darkblue, urlcolor=darkblue}

\newcommand{\infinigram}{infini-gram}
\newcommand{\Infinigram}{Infini-gram}
\newcommand{\methodname}{$\infty$-gram}
\newcommand{\ngram}{$n$-gram}
\newcommand{\ngrambold}{$\bm{n}$-gram}
\newcommand{\knn}{$k$NN}
\newcommand{\knnlm}{$k$NN-LM}

\newcommand{\retro}{\textsc{Retro}}

\newcommand{\tightparagraph}[1]{
    \vspace{-.5em} \paragraph{#1}
}
\definecolor{orange}{HTML}{E66100}
\definecolor{bluex}{HTML}{0C7BDC}
\definecolor{yellow}{HTML}{FFC20A}

\newcolumntype{L}[1]{>{\PreserveBackslash\raggedright}p{#1}}
\newcolumntype{R}[1]{>{\raggedleft\let\newline\\\arraybackslash\hspace{0pt}}m{#1}}
\newcolumntype{P}[1]{>{\centering\arraybackslash}p{#1}}

\newcommand{\ID}[1]{\cellcolor{bluex!25}\text{{#1}}}
\newcommand{\SD}[1]{\cellcolor{yellow!25}\text{{#1}}}
\newcommand{\OD}[1]{\cellcolor{orange!25}\text{{#1}}}
\newcommand{\IDRel}[1]{\cellcolor{bluex!25}$_\text{(#1\%)}$}
\newcommand{\SDRel}[1]{\cellcolor{yellow!25}$_\text{(#1\%)}$}
\newcommand{\ODRel}[1]{\cellcolor{orange!25}$_\text{(#1\%)}$}
\newcommand{\rel}[1]{$_\text{(#1\%)}$}

\newcommand{\huggingface}{\raisebox{-1.5pt}{\includegraphics[height=1.05em]{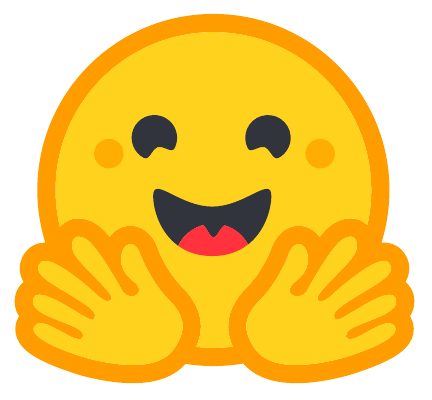}}\xspace}
\newcommand{\internet}{\raisebox{-1.5pt}{\includegraphics[height=1.05em]{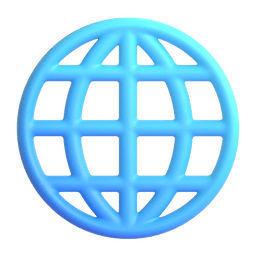}}\xspace}
\newcommand{\book}{\raisebox{-1.5pt}{\includegraphics[height=1.05em]{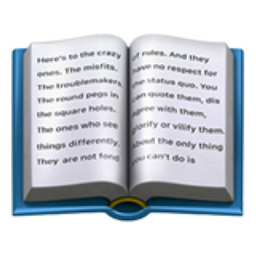}}\xspace}
\newcommand{\python}{\raisebox{-1.5pt}{\includegraphics[height=1.05em]{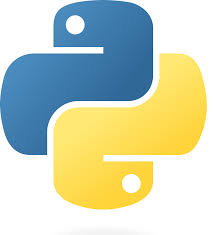}}\xspace}
\newcommand{\github}{\raisebox{-1.5pt}{\includegraphics[height=1.05em]{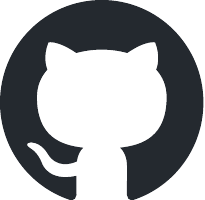}}\xspace}

\makeatletter
\newcommand*\myfontsize{%
  \@setfontsize\myfontsize{8}{9}%
}
\makeatother

\title{Infini-gram: Scaling Unbounded \ngram{} Language Models to a Trillion Tokens}

\newcommand{\aspace}{\hspace{1em}}
\newcommand{\uw}{$^{\heartsuit}$}
\newcommand{\aitwo}{$^{\spadesuit}$}
\author{%
  Jiacheng Liu\uw \aspace
  Sewon Min\uw \aspace \\
  \textbf{Luke Zettlemoyer}\uw  \aspace
  \textbf{Yejin Choi}\uw \aitwo \aspace
  \textbf{Hannaneh Hajishirzi}\uw \aitwo \aspace \\
  \uw{}Paul G. Allen School of Computer Science \& Engineering, University of Washington \\
  \aitwo{}Allen Institute for Artificial Intelligence \aspace
  \texttt{liujc@cs.washington.edu}
}

\colmfinalcopy 
\begin{document}

\maketitle

\input{sections/00-abstract}

\input{sections/01-intro}
\input{sections/02-method}
\input{sections/03-infra}
\input{sections/04-analysis}
\input{sections/05-combine}
\input{sections/06-related}
\input{sections/07-concl}

\clearpage
\ifcolmfinal
\input{sections/09-ack}
\fi

\bibliography{custom}
\bibliographystyle{colm2024_conference}

\clearpage
\appendix
\input{sections/10-app}

\end{document}

%% file: math_commands.tex
\usepackage{amsmath,amsfonts,bm}

\def\eqref#1{equation~\ref{#1}}

\def\1{\bm{1}}

\DeclareMathAlphabet{\mathsfit}{\encodingdefault}{\sfdefault}{m}{sl}
\SetMathAlphabet{\mathsfit}{bold}{\encodingdefault}{\sfdefault}{bx}{n}



%% file: sections/00-abstract.tex
\input{floats/teaser}

\begin{abstract}
Are \ngram{} language models still relevant in this era of neural large language models (LLMs)?
Our answer is \textit{yes}, and we showcase their values in both text analysis and improving neural LLMs.
This was done by modernizing \ngram{} LMs in two aspects.
First, we train them at the same data scale as neural LLMs---\textbf{5 trillion tokens}.
This is \ifcolmfinal the largest \ngram{} LM \else one of the largest \ngram{} LMs \fi ever built.
Second, existing \ngram{} LMs use small $n$ which hinders their performance; we instead allow $n$ to be arbitrarily large, by introducing a new \textbf{\methodname{} LM} with backoff.
Instead of pre-computing \ngram{} count tables (which would be very expensive), we develop an engine named \textbf{\infinigram{}}---powered by suffix arrays---that can compute \methodname{} (as well as \ngram{} with arbitrary $n$) probabilities with \textbf{millisecond-level latency}.
The \methodname{} framework and \infinigram{} engine enable us to conduct many novel and interesting analyses of human-written and machine-generated text:
we find that the \methodname{} LM has fairly high accuracy for next-token prediction (47\%), and can complement neural LLMs to greatly reduce their perplexity.
When analyzing machine-generated text, we also observe irregularities in the machine--\methodname{} agreement level with respect to the suffix length, which indicates deficiencies in neural LLM pretraining and the positional embeddings of Transformers.

\ifcolmfinal
\begin{center}
\renewcommand{\arraystretch}{1.2}
\begin{tabular}{rcl}
     \book & \textbf{Project Homepage} & \href{https://infini-gram.io/}{\path{infini-gram.io}} \\
     \huggingface & \textbf{Web Interface} & \href{https://huggingface.co/spaces/liujch1998/infini-gram}{\path{infini-gram.io/demo}} \\
     \internet & \textbf{API Endpoint} & \href{https://api.infini-gram.io/}{\path{api.infini-gram.io}} \\
     \python & \textbf{Python Package} & \href{https://pypi.org/project/infini-gram}{\path{pypi.org/project/infini-gram}} \\
     \github & \textbf{Source Code} & \href{https://github.com/liujch1998/infini-gram}{\path{github.com/liujch1998/infini-gram}}
\end{tabular}
\end{center}
\fi
\end{abstract}

%% file: floats/teaser.tex
\begin{figure*}[!b]
\centering
\vspace{-8pt}
\ifcolmfinal
\includegraphics[width=1.0 \linewidth]{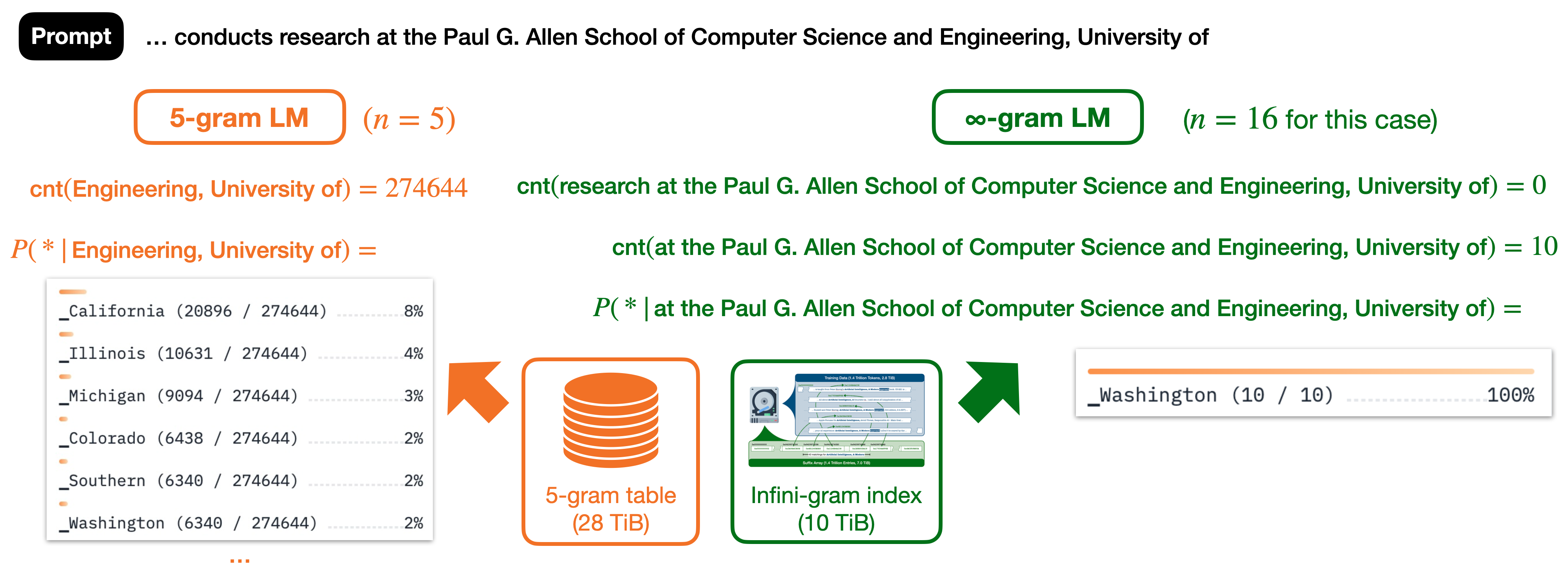}
\else
\includegraphics[width=1.0 \linewidth]{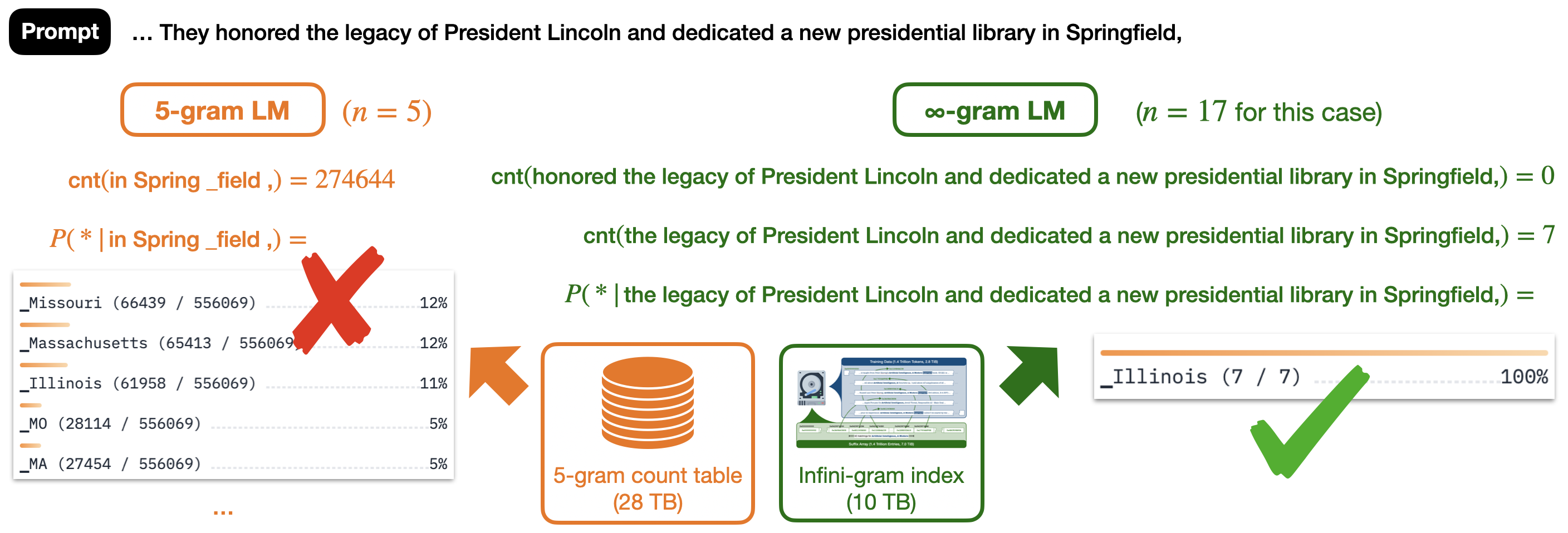}
\fi
\vspace{-12pt}
\caption{
An example where a 5-gram LM gives an incorrect prediction but the \methodname{} gives the correct prediction by using the longest suffix of the prompt that has a non-zero count in the corpus.
The counting and distribution estimate in \methodname{} LM are powered by our infini-gram engine.
}
\label{fig:teaser}
\end{figure*}

%% file: sections/01-intro.tex
\section{Introduction}
\label{sec:intro}

When pretrained on trillion-token corpora, neural large language models (LLMs) achieve groundbreaking performance \citep{touvron2023llama, openlm2023openllama, Groeneveld2023OLMo}.
However, we do not yet know how such data scale would benefit other language modeling approaches.
In particular, how well does the classical, \ngram{} language model (LM) perform if estimated from such massive corpora?
In other words, \textit{are \ngram{} LMs still relevant in this era of neural LLMs?}

Our answer is \textit{yes}.
As we will show, \ngram{} LMs are useful for both text analysis and improving neural LLMs.
Yet we need to first modernize the canonical \ngram{} LM in two aspects: the training data size, and the value of $n$.
To achieve broader data coverage, we scale up the training data for \ngram{} LMs to \textbf{5 trillion tokens}, by combining some of the largest open-source text corpora.
This is \ifcolmfinal the largest \ngram{} LM \else one of the largest \ngram{} LMs \fi ever built.
Historically, \ngram{} indexes have been built only for small $n$'s (e.g., $n \le 5$; \citet{Brants2007LargeLM}), because the size of naive \ngram{} count table grows almost exponentially wrt $n$.
We instead find significant value in increasing the value of $n$.
As illustrated in \autoref{fig:teaser}, a 5-gram LM is poorly predictive of the next token, because it discards the rich context in the prompt; meanwhile, if we can use a larger $n$ (in this case \ifcolmfinal $n = 16$\else $n = 17$\fi), the prediction becomes much more accurate.
As such, we develop our \ngram{} LM with \textit{unbounded} $n$, or in other words, an \textbf{\methodname{} LM}.
We use a variant of \textit{backoff} \citep{Jurafsky2000SpeechAL}, where we resort to smaller $n$ when longer \ngram{}s have a zero count.
Due to \textit{sparsity} in the \methodname{} estimates, in some of the later experiments (\S\ref{sec:combine}), we will interpolate between the \methodname{} LM and neural LMs to yield a hybrid LM upon which perplexity can be computed.

We develop a low-latency, resource-efficient engine to serve this massive \methodname{} LM.
Instead of building an explicit \ngram{} count table, which is infeasible for arbitrarily large $n$ and such extreme data scale, we power the \methodname{} LM with a \textit{suffix array} of the dataset -- a data structure that supports fast \ngram{} counting, and is efficient in both storage space and compute.
Our index takes 7 bytes of storage per token (3.5x overhead compared to the raw dataset), and on a dataset with 1.4 trillion tokens, it can be built with a single 128-code CPU node in about 2 days, using 10 TB of disk storage.
Average inference latency is less than \textbf{20 milliseconds} for counting an \ngram{} and finding \textit{all} positions of its occurrence (regardless of how large $n$ is or how frequently the \ngram{} appears), and under 200 milliseconds for all other query types including \methodname{} language modeling and decoding. 
All indexes stay on-disk at inference time.
We refer to this engine as \textbf{\infinigram{}}.

Analyses with \methodname{} (\S\ref{sec:analysis}) offers new insights into human-written and machine-generated text.
We found that \methodname{} has a fairly high accuracy (47\%) when predicting the next token given a prefix of a human-written document, and this accuracy is higher on tokens where the effective $n$ is larger.
In contrast, conventional \ngram{}s (with small $n$) are insufficient in capturing a long enough context to predict the next token (29\% accuracy).
Moreover, we show that \methodname{} can complement neural LMs and reach better performance when combined: heuristically interpolating between the estimates made by \methodname{} and neural LMs can greatly reduce perplexity (by up to 73\%) compared to the neural LMs alone, even when the neural LM is as large as 70B (\S\ref{sec:combine}).
When analyzing the level of agreement with \methodname{}, nucleus sampling \citep{Holtzman2019TheCC} from neural LMs produces machine-generated text with an agreement plot most similar to human-written text, among other decoding methods like greedy decoding and temperature sampling; for greedy decoding, we observe significant fluctuation in the agreement level wrt the suffix length, which indicates deficiencies in neural LM pretraining and the positional embeddings of Transformers.

We \ifcolmfinal are hosting \else will host \fi host a public web interface and an API endpoint that serve \ngram{}/\methodname{} queries on several popular open corpora: Dolma \citep{DolmaDataset}, RedPajama \citep{together2023redpajama}, Pile \citep{gao2020pile}, and C4 \citep{Raffel2019ExploringTL}.
We also release a Python package for local serving and building new indexes, as well as our source code.
We hope these tools can enable more insightful analysis and understanding of large text corpora, and open up new avenues for data-driven language modeling.

%% file: sections/02-method.tex
\section{\methodname{} LM: Extending \ngram{} LMs with Unbounded $n$}
\label{sec:method}

\tightparagraph{Background: \ngram{} LM.}
The \ngram{} LM is a classical, statistical language model based on counting the occurrences of \ngram{}s.
In its most simple form, the probability of a token $w_i$ given a context $w_{i-(n-1):i-1}$ is estimated as $P_n(w_i \mid w_{i-(n-1):i-1}) = \frac{\mathrm{cnt}(w_{i-(n-1):i-1} w_i \mid \mathcal{D})}{\mathrm{cnt}(w_{i-(n-1):i-1} \mid \mathcal{D})}$, where $\mathrm{cnt}(\mathbf{w} \mid \mathcal{D})$ is the number of times the $n$-gram $\mathbf{w}$ appears in the training data $\mathcal{D}$ (i.e., a corpus), and $n$ is a pre-defined hyperparameter.
(When $n = 1$, we define $w_{i-(n-1):i-1}$ as the empty string $\varepsilon$, whose count is equal to $|\mathcal{D}|$.)
However, this naive version of \ngram{} LM faces the sparsity issue: the numerator may be zero, resulting in an infinite perplexity.
One common solution is \textbf{backoff} \citep{Jurafsky2000SpeechAL}: on an instance-wise basis, when the numerator is zero we decrease $n$ by one, and do this repeatedly until the numerator becomes positive.
One caveat in backoff is that it does not yield a valid distribution for $P_n(* | w_{i-(n-1):i-1})$, because the \textbf{effective $\mathbf{n}$} is dependent on $w_i$.
Therefore, further probability discounting is required to normalize this distribution (e.g., Katz backoff \citep{Katz1987EstimationOP}).

Conventionally, \ngram{} LMs have been implemented by building an \ngram{} count table of the training data.
This table stores all unique \ngram{}s that appear in the training data, each associated with its count.
Such \ngram{} count tables are huge and grow almost exponentially wrt $n$.
For example, the 5-gram count table for a 1.4-trillion-token corpus would consume 28 TB of disk space.
As a result, previous \ngram{} LMs are limited to very small $n$, most commonly $n = 5$, and to frequent \ngram{}s only (e.g., \citet{franz2006all}).
As we illustrated in \autoref{fig:teaser} and will further quantify in \S\ref{sec:analysis}, the problem with small $n$ is that it discards richer context, making such \ngram{} LMs poorly predictive of future tokens.

\tightparagraph{\methodname{} LM.}
The \methodname{} LM is a generalization of the \ngram{} LM, where conceptually we start backing off from $n = \infty$.
We use a variant of backoff: we backoff only when the denominator is zero.
This means we stop backing off as soon as the denominator becomes positive, upon which the numerator might still be zero.
On an instance-wise basis, the effective $n$ is equal to one plus the length of the prompt's \textbf{longest suffix} that appears in the training data.

\textbf{For the rest of this paper, we will use ``\methodname{}'' to refer to the \methodname{} LM.}
\methodname{} is formally defined as
\begin{align*}
P_\infty(w_i \mid w_{1:i-1})
&= \frac{\mathrm{cnt}(w_{i-(n-1):i-1} w_i \mid \mathcal{D})}{\mathrm{cnt}(w_{i-(n-1):i-1} \mid \mathcal{D})}
\end{align*}
where $w_{1:i-1}$ are all tokens preceding $w_i$ in the document, and
\begin{align*}
n &= \max \{ n' \in [1, i] \mid \mathrm{cnt}(w_{i-(n'-1):i-1} \mid \mathcal{D}) > 0 \}.
\end{align*}
Unlike Katz backoff, $P_\infty(* | w_{1:i-1})$ is a valid distribution by construction and does not require discounting.
This is because the effective $n$ is solely dependent on $w_{1:i-1}$ and does not depend on $w_i$, and $\sum_{w_i \in \mathcal{V}} \mathrm{cnt}(w_{i-(n-1):i-1} w_i \mid \mathcal{D}) = \mathrm{cnt}(w_{i-(n-1):i-1} \mid \mathcal{D})$.

Further, we define the \textbf{sparsity} of this \methodname{} estimate: an estimate is sparse iff $P(w_i | w_{i-(n-1):i-1}) = 1$ for one of the $w_i \in \mathcal{V}$, and is zero for all other tokens in the vocabulary.
Intuitively, this means there is only one possible next token given this context, according to the training data.
As we will show in \S\ref{sec:analysis}, sparse estimates are more predictive of the actual tokens than non-sparse ones.

\tightparagraph{Interpolating with neural LMs.}
\methodname{} estimates contain zero probabilities, which may lead to infinite perplexity.
We do \textit{not} attempt to compute the perplexity of the \methodname{} itself.
Instead, we interpolate it with neural LMs and show perplexity improvement over the neural LMs alone (\S\ref{sec:combine}).
The combined model is
\begin{align*}
    P(y \mid x) = \lambda P_\infty(y \mid x) + (1-\lambda) P_\text{neural}(y \mid x),
\end{align*}
where $\lambda \in [0, 1]$ is a hyperparameter.

%% file: sections/03-infra.tex
\section{\Infinigram{}: A Performant Engine for \ngram{}/\methodname{} Queries}
\label{sec:infra}

We train \methodname{} on modern, trillion-token text corpora.
However, it is practically infeasible to build \ngram{} count tables with unbounded $n$ for such massive datasets. 
In this section, we describe our \infinigram{} engine that processes \ngram{}/\methodname{} queries efficiently.
\Infinigram{} is powered by a data structure called \textbf{suffix array}.
We will show how to build this suffix array index and how to perform \ngram{}/\methodname{} inferences with it.

\input{floats/sa}

\tightparagraph{Suffix array.}
The essence of \ngram{} and \methodname{} LMs is counting a given \ngram{} in the training data.
As such, we leverage the suffix array data structure, which is originally designed for efficiently counting the number of times a given ``needle'' string (length $L$) appears as substring of a huge ``haystack'' string (length $N$).
When the suffix array is built for a haystack string, counting a given needle string has time complexity $O(L + \log N)$.

A suffix array represents the lexicographical ordering of all suffixes of an array (or a string, which is an array of characters).
For an array of length $N$, the suffix array contains $N$ unique integers, where the $i$-th element is the starting position of the suffix ranked $i$-th among all suffixes.
\autoref{fig:sa} (left) shows the suffix array for a toy string, \texttt{aabaca}.

As shown in \autoref{fig:sa} (right), we build the suffix array on the byte array of the tokenized dataset (i.e., \textbf{token array}).
Documents are separated by the \texttt{\textbackslash{}xff\textbackslash{}xff} token.
In the token array, each consecutive two bytes represent a token ID (assuming that $|\mathcal{V}| < 2^{16} = 65536)$.
Given that the dataset has $N$ tokens, the token array has $2 N$ bytes.
The suffix array contains $N$ elements, each pointing to a token in the token array by storing its byte offset.
Every tokens in the token array appears exactly once in the suffix array.
Each pointer can be stored with $\lceil \log_2 (2 N) / 8 \rceil$ bytes.
For corpora with 2B to 500B tokens (which is the range we deal with, after sharding (\S\ref{subapp:indexing})), this is 5 bytes per pointer, and thus the suffix array has $5 N$ bytes.
Therefore, the combined size of token array and suffix array (i.e., the \textbf{\infinigram{} index}) is $7 N$ bytes.

\tightparagraph{Building the suffix array.}
Suffix arrays can be built in linear time with respect to the size of the token array \citep{Krkkinen2006LinearWS}.
We adapted from the suffix array implementation in \citet{lee2021deduplicating} and further optimized it for efficiency.
It took us $\sim$48 hours to build the suffix array for RedPajama on a single node with 128 CPUs and 1TiB RAM.
We have built the suffix arrays for Dolma (3T tokens), RedPajama (1.4T tokens), Pile (380B tokens), and C4 (200B tokens).
Since \infinigram{} indexes are additive (\S\ref{subapp:sa}), together these can be easily combined into a larger index with a total of \textbf{5 trillion tokens}, and the implicit count table contains at least \textbf{2 quadrillion unique \ngram{}s} (\S\ref{subapp:table-size}).

\tightparagraph{Inference with the suffix array.}
Computing the \ngram{} LM probability involves counting the number of occurrences of a token string, i.e., $\mathrm{cnt}(x_1 ... x_n)$.
By construction, the occurrence positions of strings starting with $x_1 ... x_n$ lies in a single, consecutive segment in the suffix array.
Thus we only need to find the first and last occurrence positions, and the count would be the difference between them.
Beyond counting and \ngram{}/\methodname{} language modeling, \infinigram{} can also be used to retrieve documents containing an \ngram{}, or a CNF expression with multiple \ngram{}s (\S\ref{subapp:inference}).

During inference, the entire \infinigram{} index can stay on-disk, which minimizes the compute resources needed (no GPU, and minimal CPU / RAM).
In \S\ref{subapp:inference}, we discuss several optimization techniques applied to the inference engine: parallelized shard processing, hinted search, memory pre-fetching, fast effective-$n$ lookup, and amortized query processing.
On RedPajama, our most optimized \infinigram{} engine can count a given \ngram{} with an average latency of less than 20 milliseconds.
It can compute the probability and next-token distribution in 40 milliseconds for \ngram{} LMs, and in 200 milliseconds for the \methodname{}.
See \S\ref{subapp:query_types} for the full list of supported query types and additional details on latency benchmarking.

%% file: floats/sa.tex
\begin{figure*}[!t]
\centering
\includegraphics[width=0.9 \textwidth]{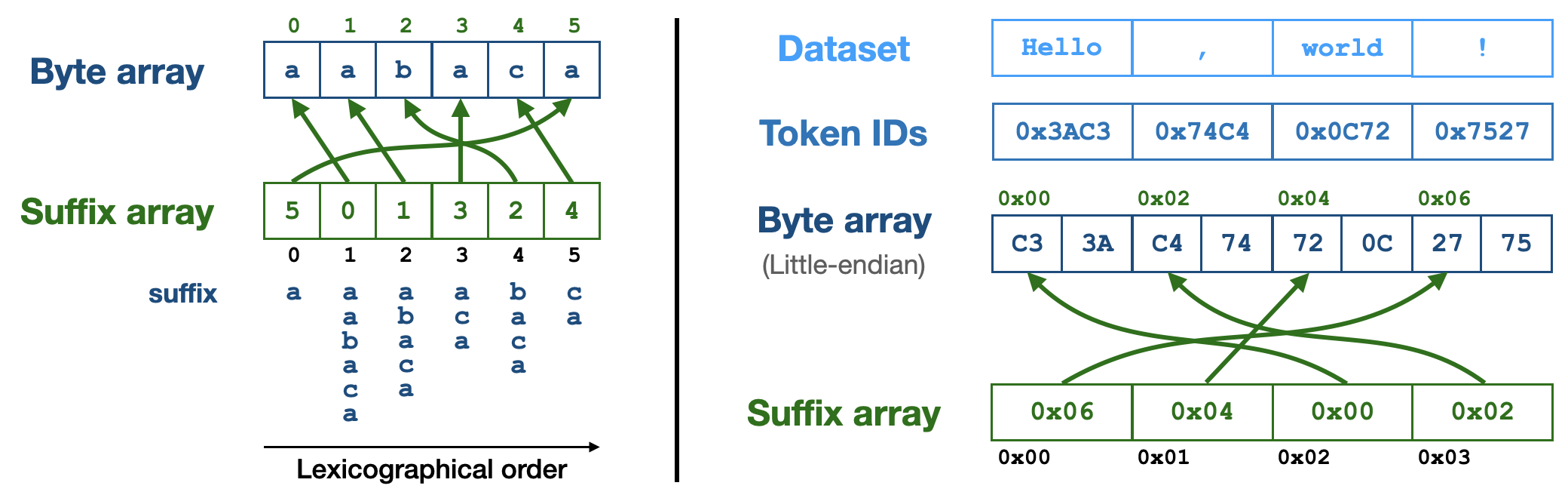}
\vspace{-5pt}
\caption{
    \textbf{Left:} the suffix array for a toy string.
    \textbf{Right:} illustration of the suffix array in the \infinigram{} index, with $N = 4$ tokens in the dataset.
}
\vspace{-12pt}
\label{fig:sa}
\end{figure*}

%% file: sections/04-analysis.tex
\section{Analyzing Human-written and Machine-generated Text using \methodname{}}
\label{sec:analysis}

In this section, we present some analyses of human-written and machine-generated text from the perspective of \methodname{}, mostly focusing on the token-wise agreement between \methodname{}'s prediction and the actual text.
In summary, we found that:
\begin{enumerate}[itemsep=2pt, leftmargin=16pt, topsep=0pt]
\item \methodname{} has a fairly high accuracy (47\%) when predicting the next token given a prefix of a human-written document, and this accuracy is higher when a longer suffix of the prompt can be used (i.e., when the effective $n$ is larger);
\item Conventional \ngram{} LMs ($n \le 5$) are insufficient for capturing a long enough context to determine the next token, while our \methodname{} method is highly predictive of human-written and machine-generated text;
\item \methodname{} has significant potential to complement and improve neural LMs when predicting human-written text (which we further investigate in \S\ref{sec:combine});
\item When plotting the agreement level with respect to the suffix length, text generated by neural LMs with nucleus sampling is most similar to human-written text, among other decoding methods like greedy decoding and temperature sampling. For greedy decoding, the agreement plot suffers from significant fluctuation, which may be rooted in deficiencies in neural LM pretraining and the positional embeddings of Transformers.
\end{enumerate}

\tightparagraph{\methodname{} training data.}
For analyses in this section, we use a decontaminated version of Pile's training set (``Pile-train'') \citep{gao2020pile} as training data for the \methodname{}.
We built an \infinigram{} index on Pile-train using the Llama-2 tokenizer \citep{touvron2023llamatwo}, and yield 360 billion tokens.

\tightparagraph{Decontamination.}
It is important that the training data is decontaminated against the evaluation data, because otherwise it is very easy for \methodname{} to cheat by copying from the very same document in the training data.
We run decontamination of Pile-train against its validation and test sets (``Pile-val'' and ``Pile-test'', which we will use for evaluation below and also in \S\ref{sec:combine}), using the method from \citet{bff} that filters out a document if it has excessive \ngram{} overlap with the evaluation data.
See \S\ref{app:decontamination} for more details.

Decontamination is non-trivial, and its definition could vary (e.g., when there is an identical sentence, is it contamination, or is it a quote that naturally occurs in real test-time scenarios?)
Thus we followed the standard best practices for decontamination. 

\subsection{Human-written text}

\tightparagraph{Setup.}
We use Pile-val set as the human-written text.
For this analysis, we sampled 50 documents from each domain of Pile-val, and truncated each document to 1024 tokens (so the total number of tokens per domain is about 50k).
We aggregate results from all domains.

We measure the token-wise \textit{agreement} between \methodname{}'s prediction and the actual human-written text.
Since computing the full next-token distribution (or the argmax of it) in \methodname{} is relatively slow, we compute the \methodname{} probability of the actual next-token, and deem it as accurate if this probability is higher than 0.5.
(This is a lower-bound of argmax accuracy, though the gap is small.)
We further categorize all tokens by their effective $n$, i.e., one plus the length of their prompt's longest suffix that has a non-zero count in the training data.
For each category, we visualize the number of such tokens (in gray bars) as well as the agreement level (in green dots) in the middle plot of \autoref{fig:analysis_human_ngram}.

\input{floats/analysis_human_ngram}

\tightparagraph{Results.}
Overall, \methodname{} agrees with the human-written text on 47\% of the tokens.
We see that \methodname{} becomes more accurate with the increase of effective $n$: when the effective $n$ $\ge 16$, agreement is higher than 75\%.
Further analysis (Appendix \autoref{fig:analysis_human_ngram_more}) shows that the count of this longest suffix in the training data does not affect agreement substantially.

In the left plot of \autoref{fig:analysis_human_ngram}, we show the same analysis for a 5-gram LM trained on the same data, and it has much lower agreement than the \methodname{}.
5-gram LMs, which has been used extensively in previous literature \citep{franz2006all, Aiden2011QuantitativeAO}, does not capture a long enough context to correctly predict the next token: over 90\% tokens in the evaluation data has an effective $n$ of at least 5, and the \methodname{} analysis shows that the median of effective $n$ is 7 (and mean is 9.1).

In the right plot of \autoref{fig:analysis_human_ngram}, we show the same analysis for only tokens with a sparse \methodname{} estimate, which covers more than 50\% of all tokens.
The overall agreement is even higher (75\%), and when the effective $n$ $\ge 14$, agreement is higher than 80\%.
This means when the next token is unique according to the training data, that unique token is very likely to be the actual token in human-written text.

Qualitatively, we found that \methodname{} is often good at completing multi-token words (e.g., \texttt{hippop\underline{otamus}}, correctly predicted tokens are underlined), common phrases (e.g., \texttt{born \underline{in}}), and entity names (e.g., \texttt{educated at Tr\underline{inity College}}).
\methodname{} is not very good at recalling factual knowledge (e.g., predicting the first token of an entity name), likely due to insufficient contextualization.

\input{floats/analysis_human_neural_ngram}

\tightparagraph{\methodname{} can shine where neural LMs fail.}
In \autoref{fig:analysis_human_neural_ngram}, we plot the distribution of probabilities assigned by the Llama-2 70B/13B/7B models \citep{touvron2023llamatwo} to the actual tokens in human-written text, and the human--\methodname{} agreement for tokens in each probability bucket.
(The higher the assigned probability, the higher agreement Llama-2 has with the actual tokens.)
We observe a positive, yet imperfect, correlation between neural LMs and \methodname{} regarding their agreement with the actual text.
In particular, when the neural LM performance is very poor (left side of the histogram), \methodname{} still gives a non-trivial agreement of above 20\%; if only considering tokens with sparse \methodname{} estimates, the agreement is as high as 50\%.
This indicates a huge potential of complementing and improving the performance of neural LMs with \methodname{} for predicting human-written text, which we further investigate in \S\ref{sec:combine}.

\subsection{Machine-generated text}

\tightparagraph{Setup.}
Similar to the analysis with human-written text, we sampled 50 documents from each domain of Pile-val.
We use the first 50 tokens of each document to prompt neural LMs to generate a continuation.
Generation continues up to the original length of the document, or when an [EOS] token is generated.
We experiment with three decoding methods: greedy decoding, temperature sampling, and nucleus sampling \citep{Holtzman2019TheCC}.
The neural LMs are Llama-2 70B/13B/7B, GPT-J 6B \citep{gpt-j}, and GPT-Neo 2.7B/1.3B/125M \citep{gao2020pile}.
The tokenizer of GPT-Neo/J is different from Llama-2, so we built a separate version of \infinigram{} index for Pile-train based on the GPT-Neo/J tokenizer.

\input{floats/analysis_machine_ngram}

\tightparagraph{Impact of decoding method.}
The top row of \autoref{fig:analysis_machine_ngram} shows results of the three decoding method on the same neural LM -- Llama-2 70B.
In general, increasing stochasticity shifts the effective $n$ to the smaller side, and also decreases the agreement level.
Nucleus sampling has the most similar distribution of effective $n$ compared to human-written text (\autoref{fig:analysis_human_ngram}, middle plot), which is probably why nucleus sampling is usually preferred in text generation.
Greedy decoding has even higher effective $n$ than human-written text, which implies that greedy decoding could lead to over-memorization of training data as well as lack of diversity.

\tightparagraph{Impact of model size.}
The bottom row of \autoref{fig:analysis_machine_ngram} shows the same analysis for different sizes of neural LM under greedy decoding.
In general, increasing model size slightly increases the effective $n$, and also increases the agreement level.
This indicates that larger models memorizes more from the training data, and are also more inclined to copy verbatim.
The agreement level of GPT-Neo/J models is higher than Llama-2 models, probably because GPT-Neo/J are trained on the same data as the \methodname{} (i.e., Pile-train).
Overall, text generated by these neural LMs has similar agreement level with \methodname{} as human text.

One very curious phenomenon is that, as effective $n$ increases, the agreement level fluctuates greatly in greedy decoding (but not nucleus or temperature sampling, where agreement level almost increases monotonically).
Such fluctuation is even more rapid for smaller models (Llama-2 13B/7B and GPT-Neo/J models), and for Llama-2 7B the fluctuation is even periodic (rapidly dropping at effective $n = 20, 24, 28, 32$; this is statistically significant, a two-proportion z-test gives a p-value of $< 10^{-99}$).
We suspect that this may be caused by the application of positional embeddings when pretraining these Transformer-based models, and we welcome further investigation from the community.

%% file: floats/analysis_human_ngram.tex
\begin{figure*}[!t]
\begin{minipage}{0.325 \textwidth}
\includegraphics[width=\textwidth]{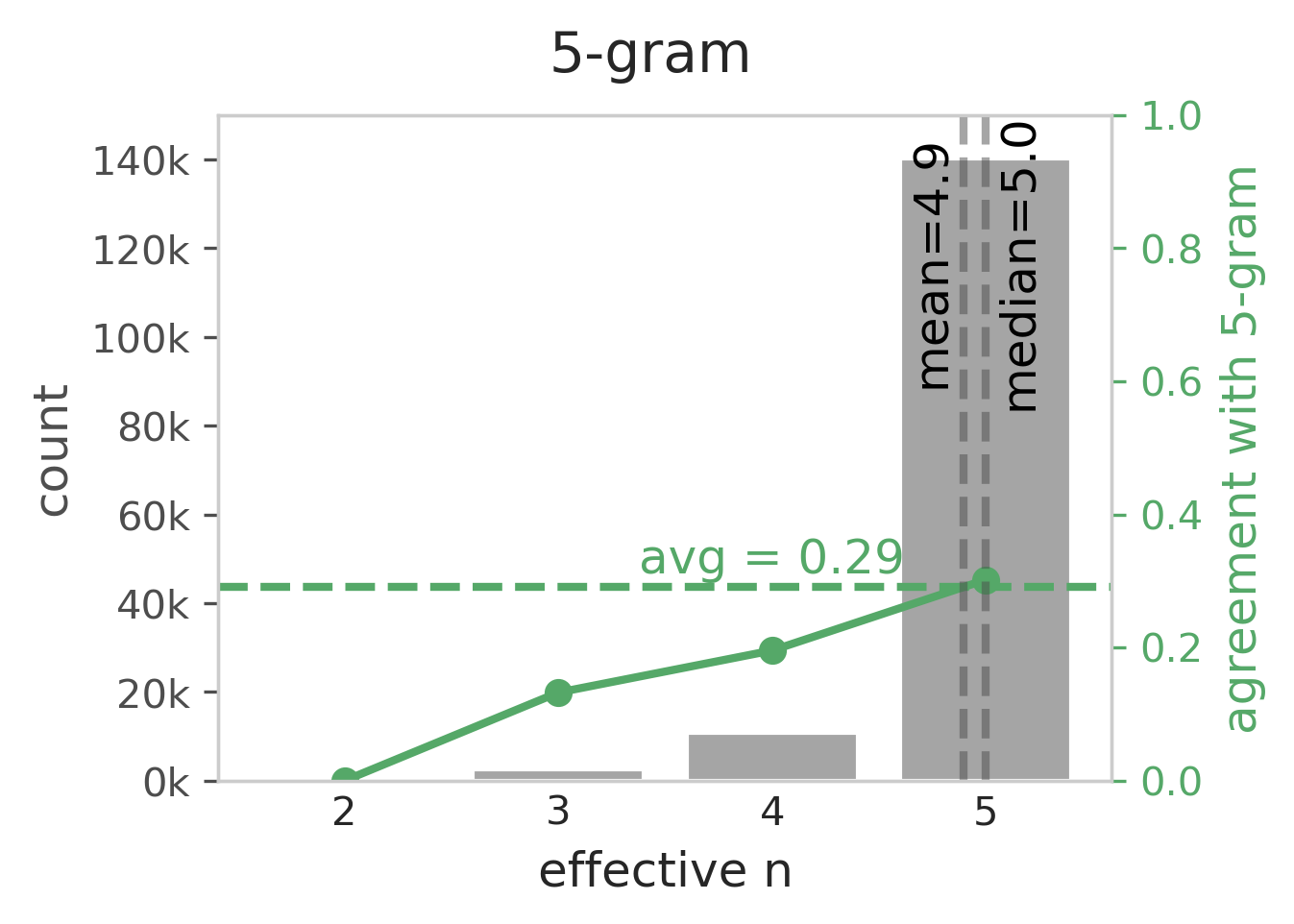}
\end{minipage}
\hfill
\begin{minipage}{0.325 \textwidth}
\includegraphics[width=\textwidth]{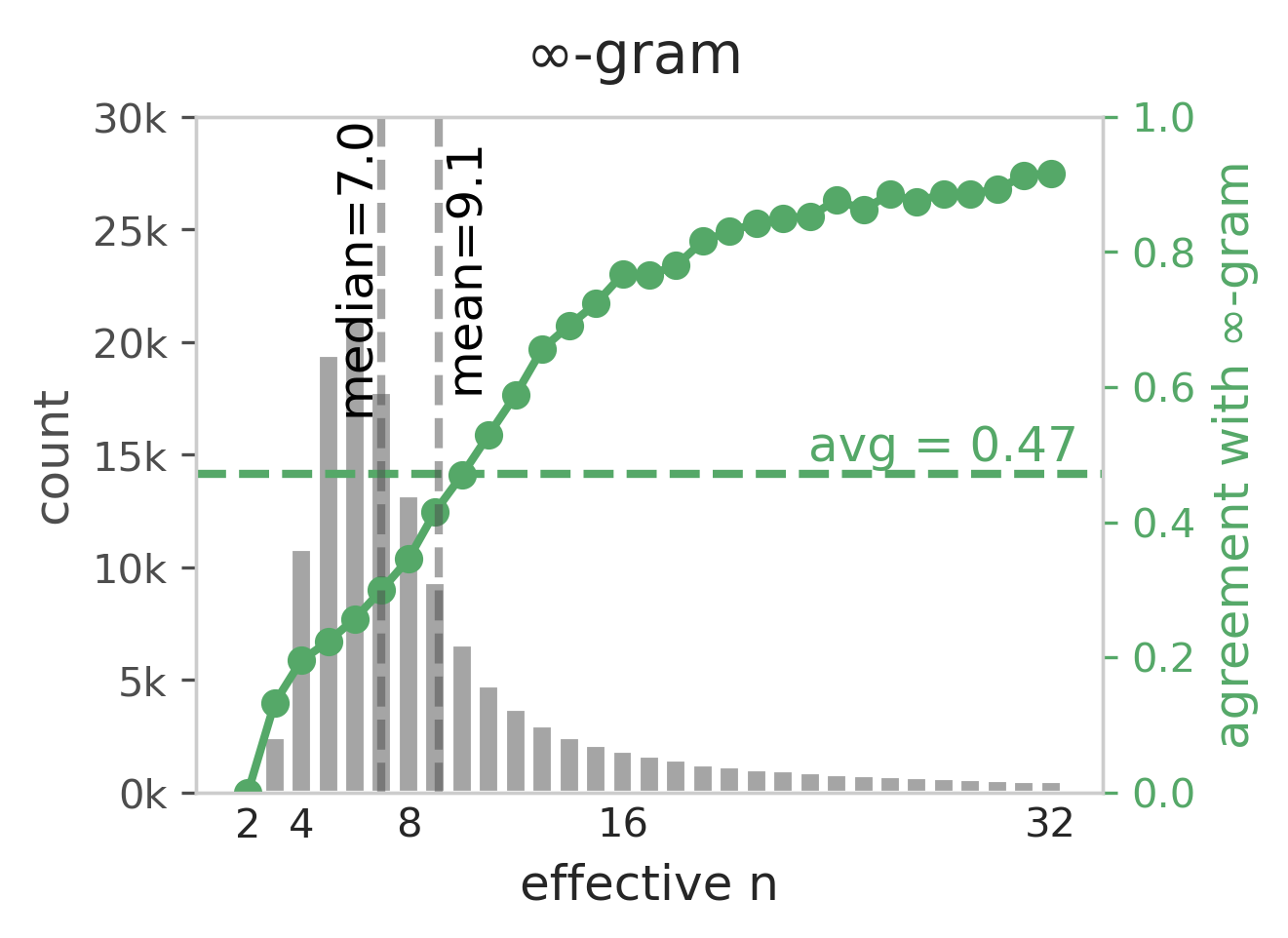}
\end{minipage}
\hfill
\begin{minipage}{0.325 \textwidth}
\includegraphics[width=\textwidth]{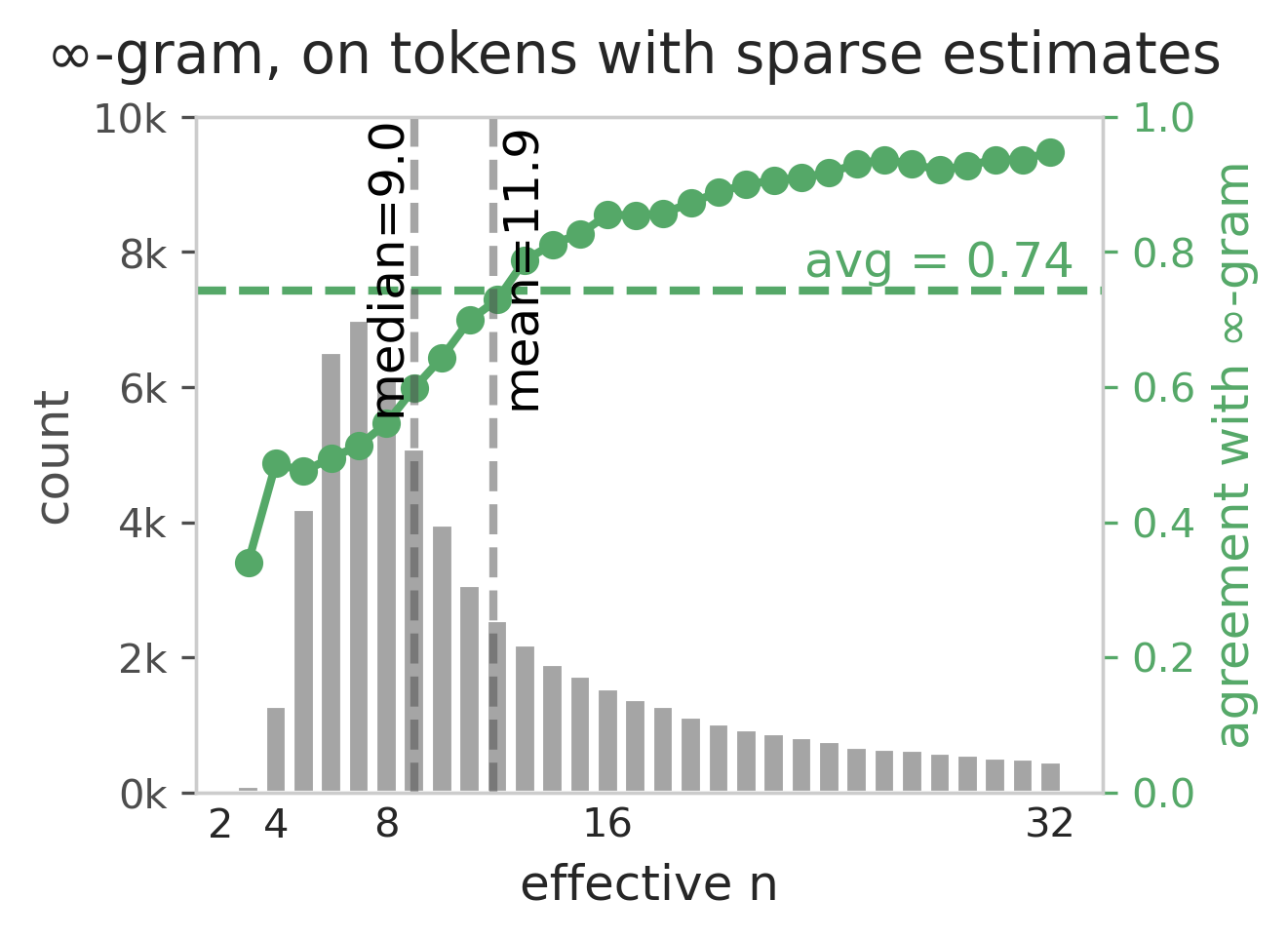}
\end{minipage}
\vspace{-8pt}
\caption{
    Token-wise agreement between human-written text and \ngram{}/\methodname{} LMs.
}
\vspace{-8pt}
\label{fig:analysis_human_ngram}
\end{figure*}

%% file: floats/analysis_human_neural_ngram.tex
\begin{figure*}[!t]
\centering
\includegraphics[width=0.40 \textwidth]{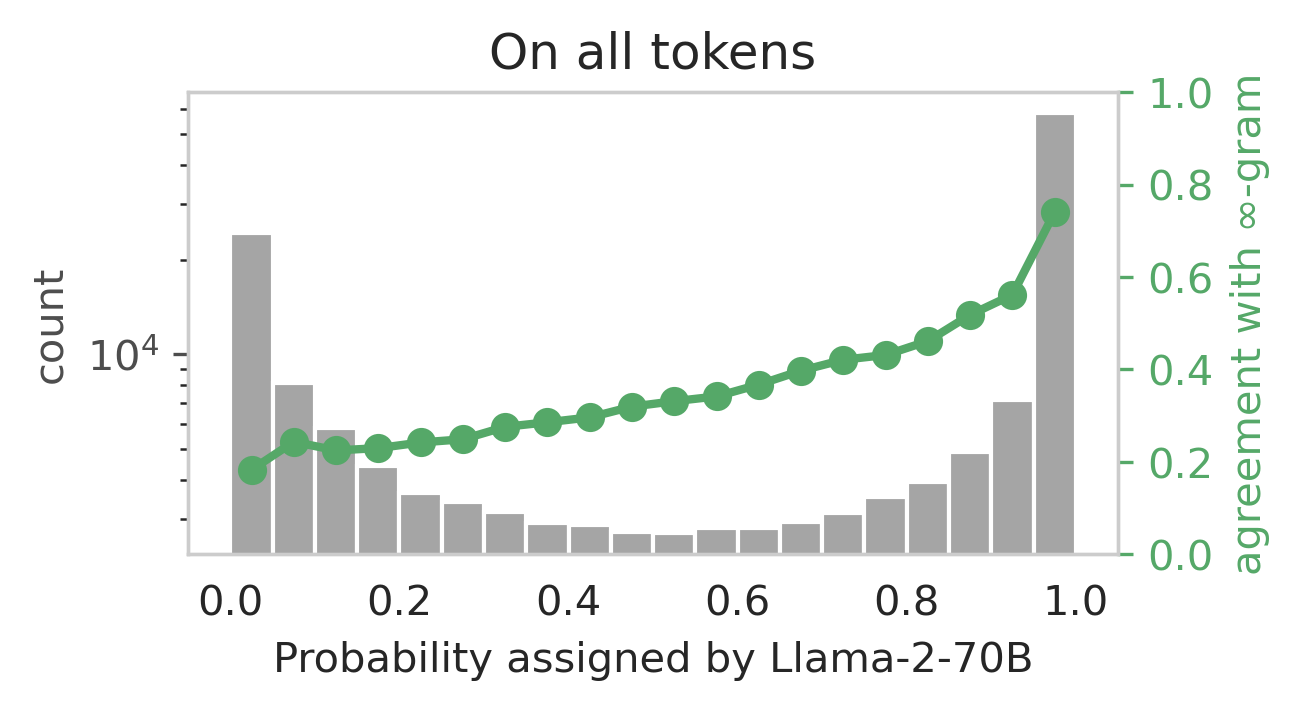}
\includegraphics[width=0.40 \textwidth]{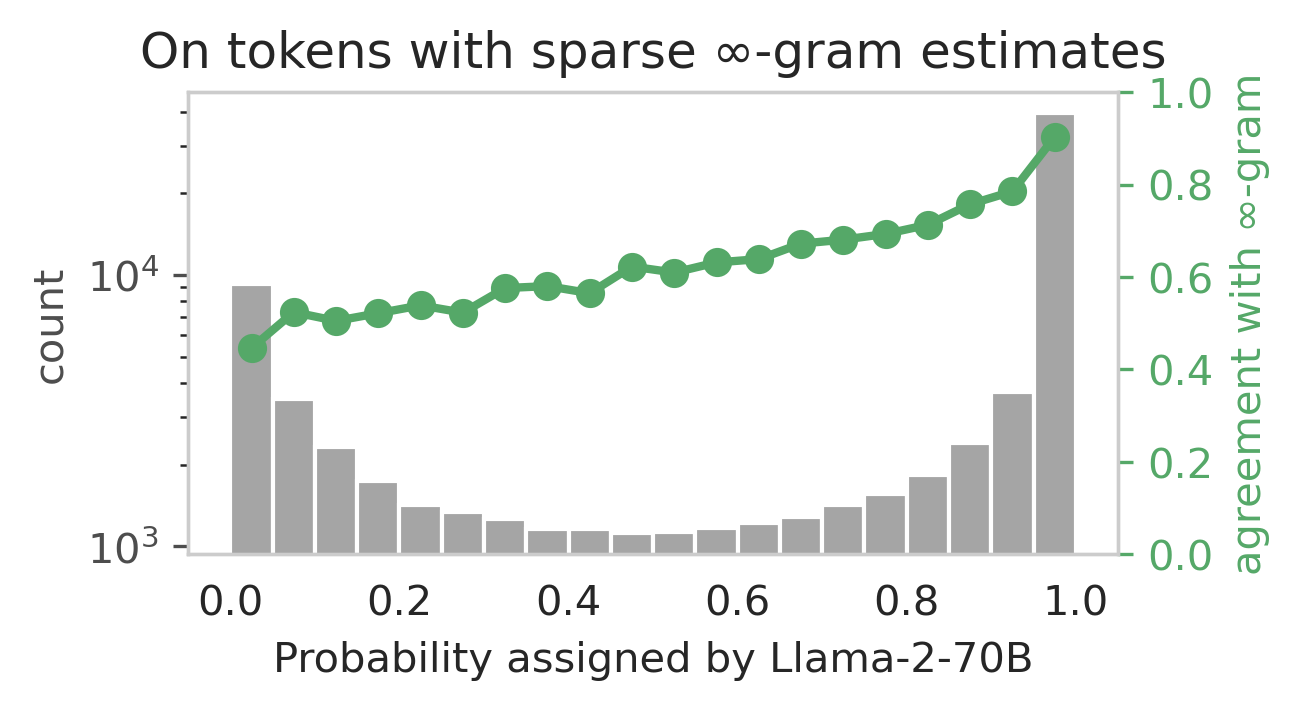}
\vspace{-6pt}
\caption{
    Distribution of probabilities assigned by neural LMs to human-written text tokens, and \methodname{}'s agreement with these tokens.
    \textbf{Takeaway:} \methodname{} and neural LMs are predictive of actual human text on different tokens, and thus \methodname{} estimates -- especially sparse \methodname{} estimates -- can be used to complement neural LMs.
    See \autoref{fig:analysis_human_neural_ngram_more} for extended results on Llama-2 13B/7B models.
}
\vspace{-10pt}
\label{fig:analysis_human_neural_ngram}
\end{figure*}

%% file: floats/analysis_machine_ngram.tex
\begin{figure*}[!t]
\includegraphics[width=0.325 \textwidth]{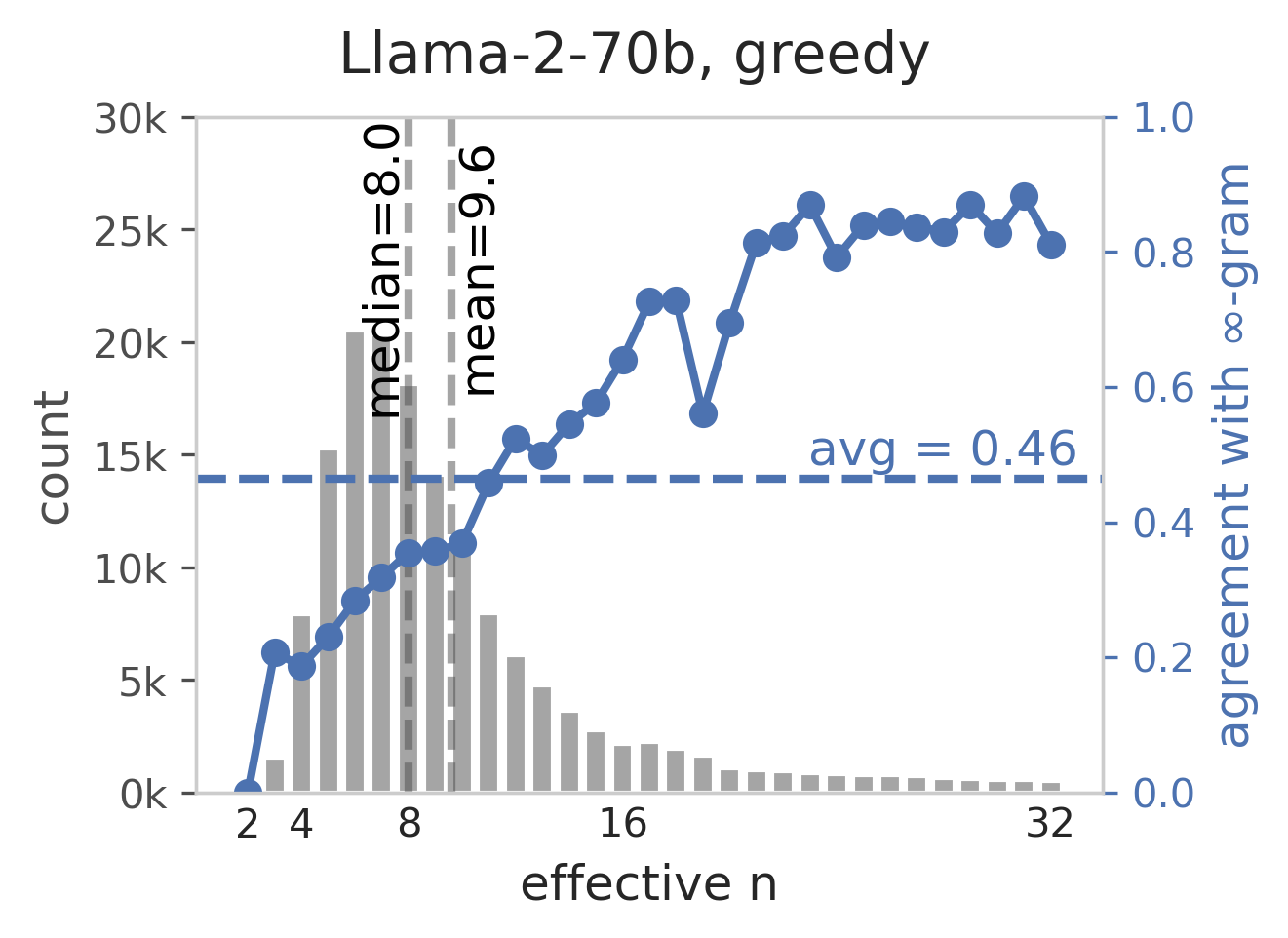}
\includegraphics[width=0.325 \textwidth]{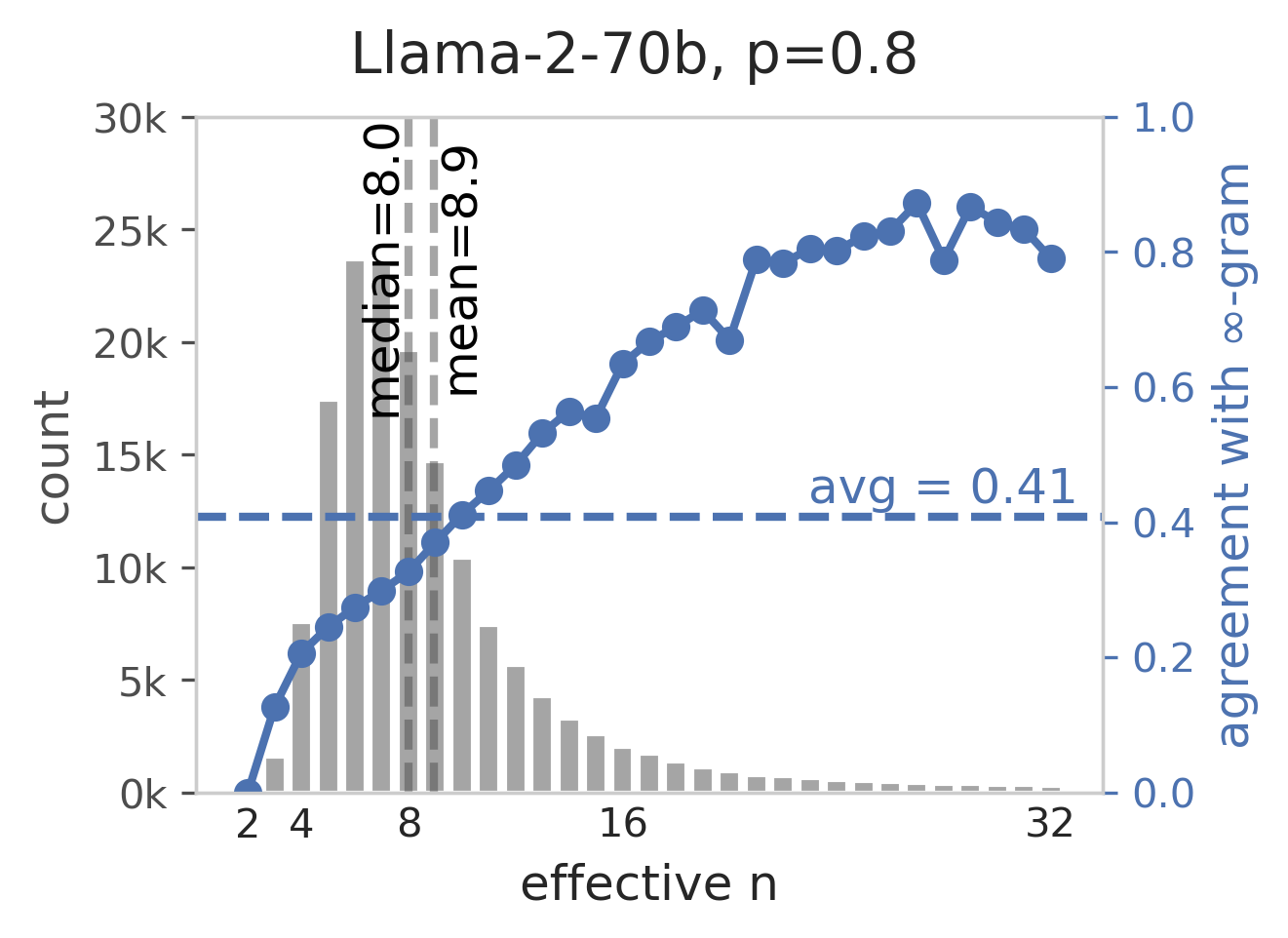}
\includegraphics[width=0.325 \textwidth]{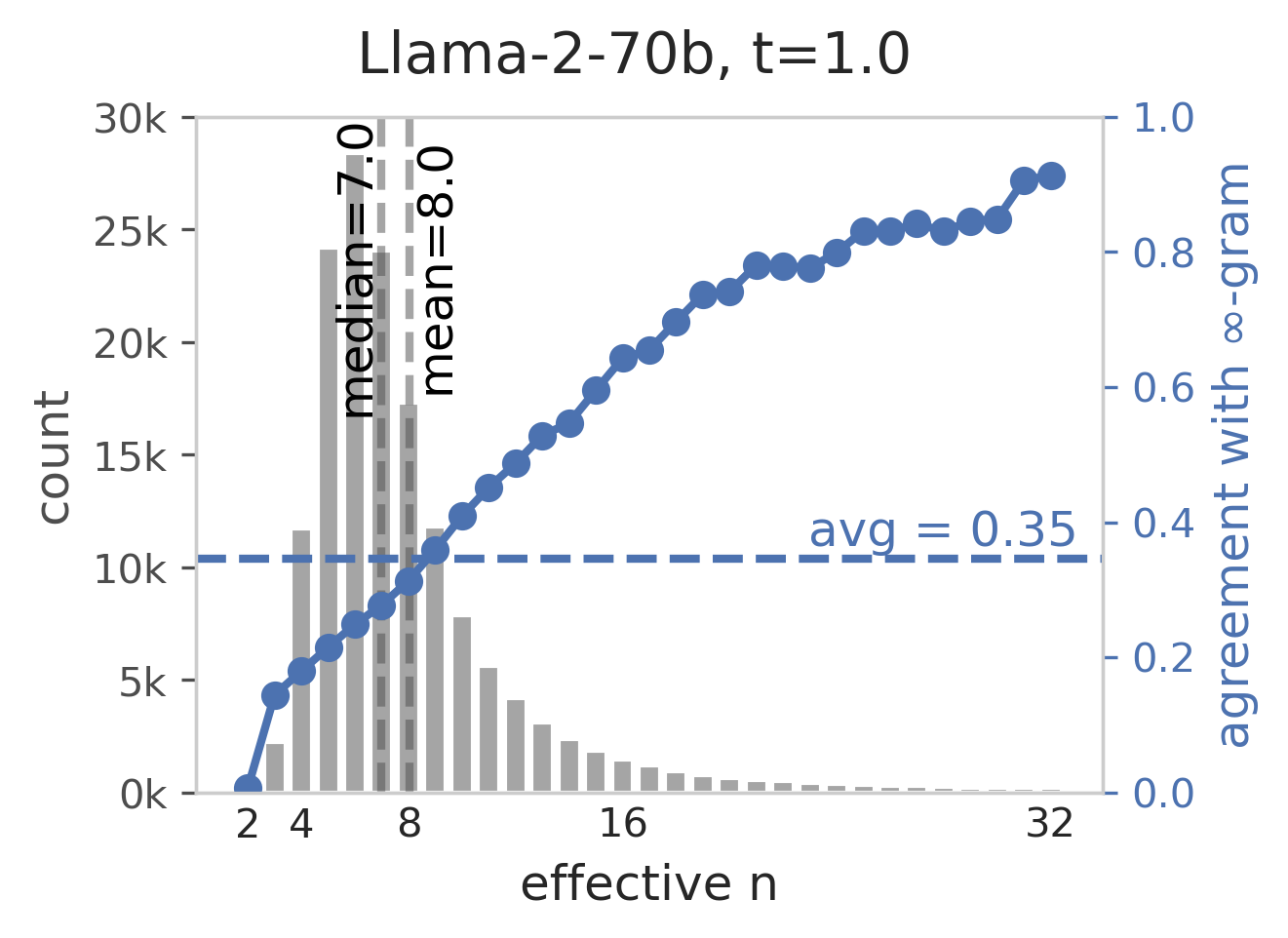}
\includegraphics[width=0.325 \textwidth]{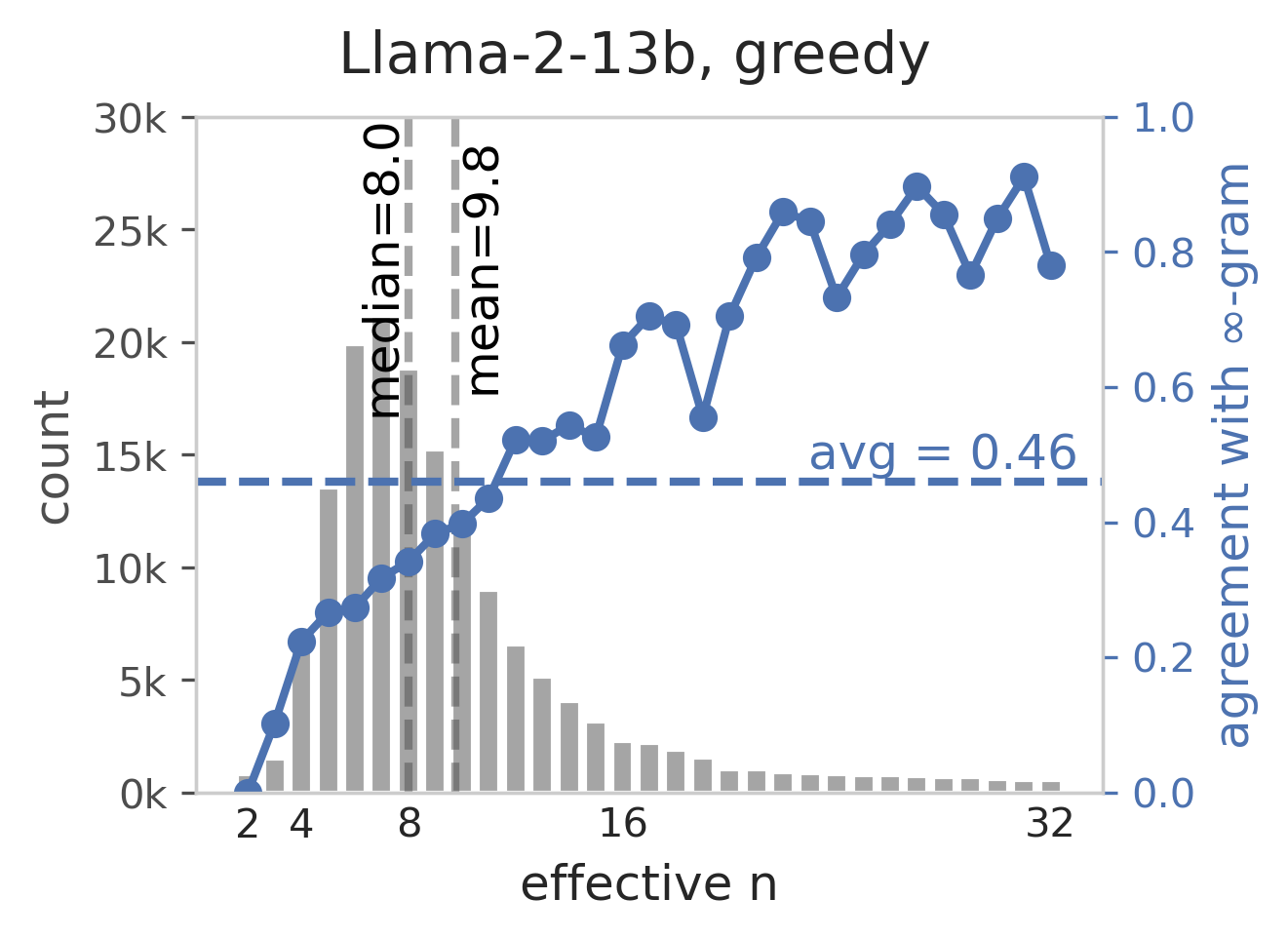}
\includegraphics[width=0.325 \textwidth]{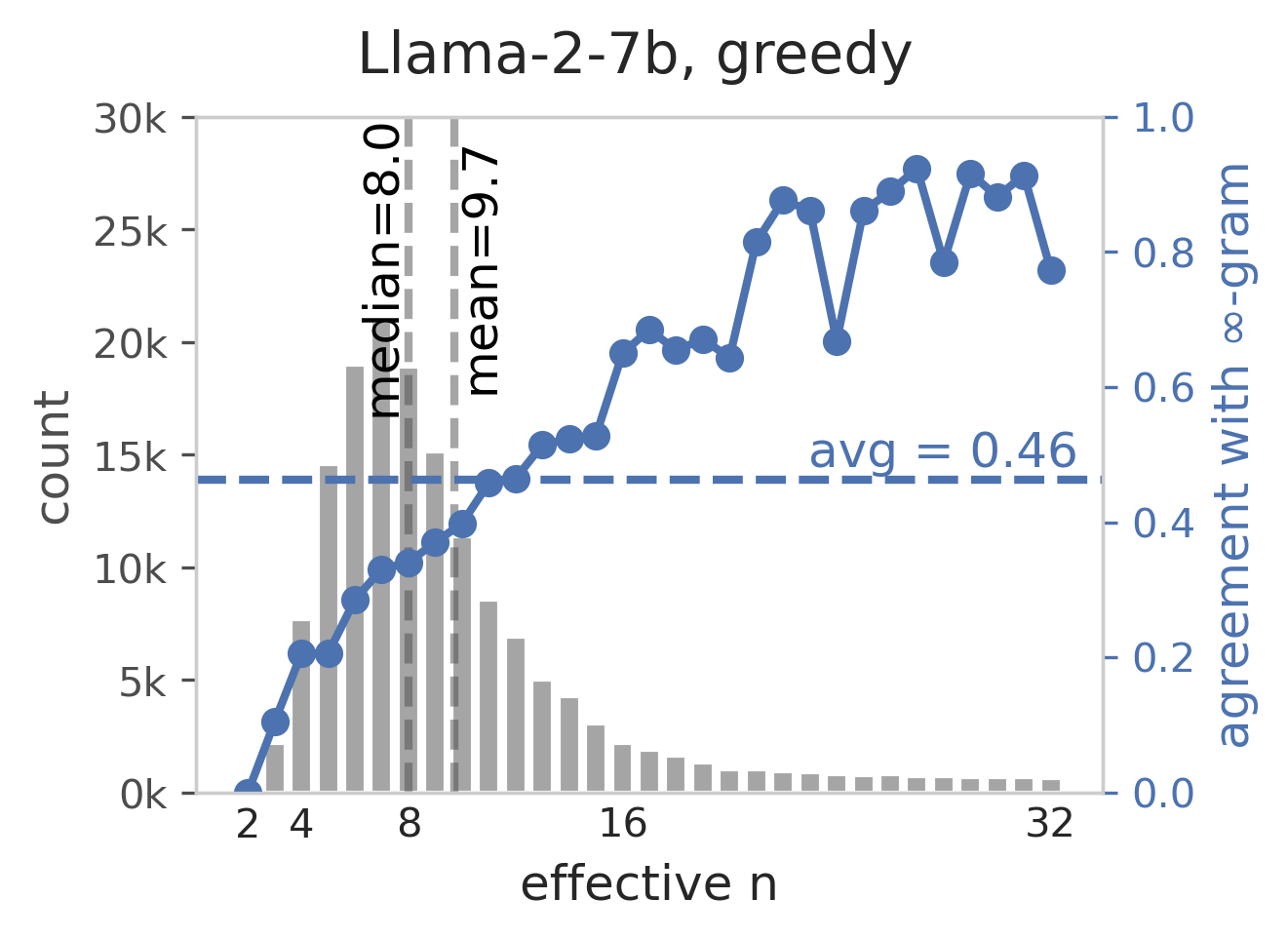}
\includegraphics[width=0.325 \textwidth]{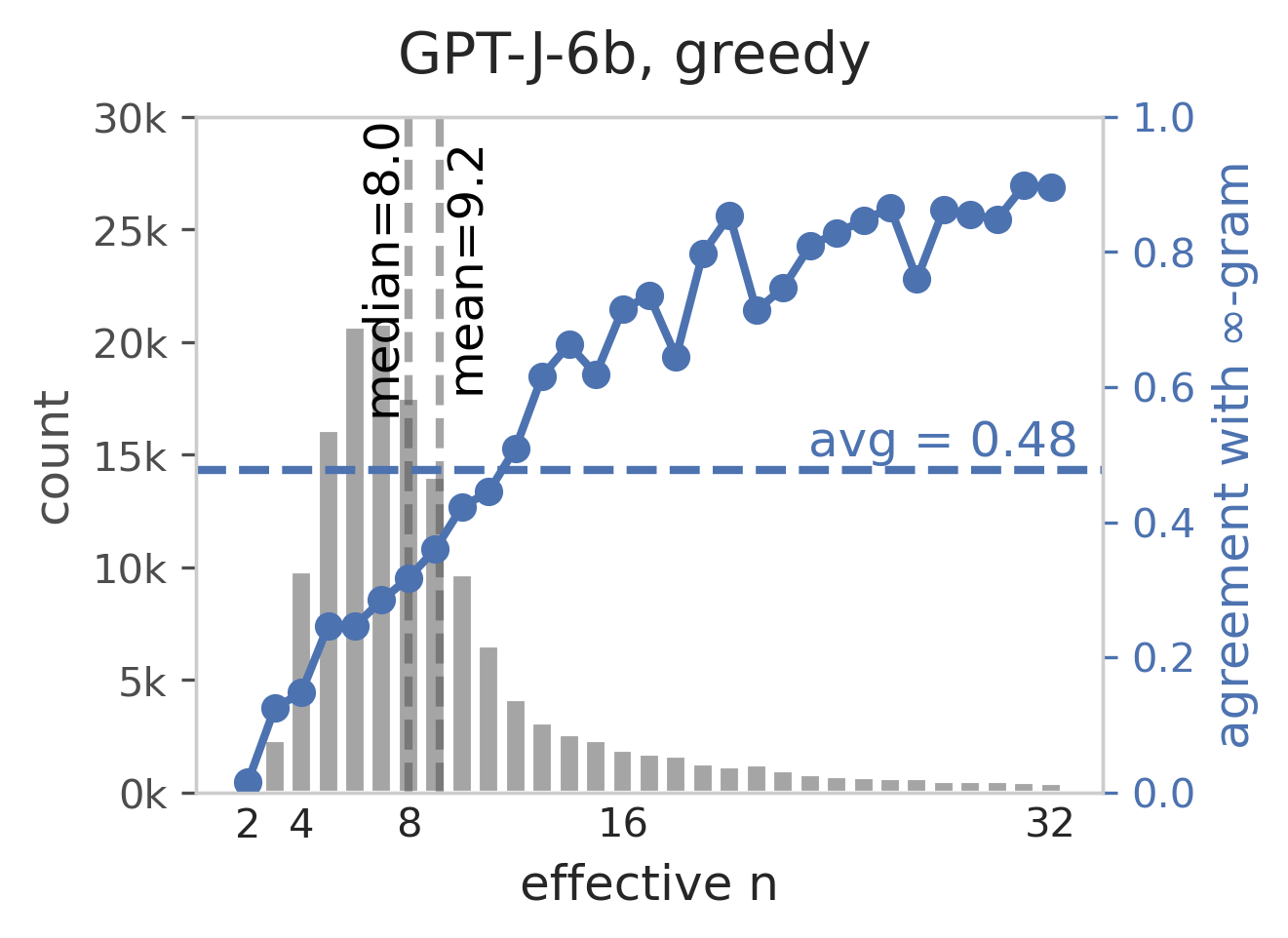}
\vspace{-8pt}
\caption{
    Token-wise agreement between machine-generated text and \methodname{}.
    All tokens are considered.
    See \autoref{fig:analysis_machine_ngram_more} for results on GPT-Neo models.
}
\vspace{-12pt}
\label{fig:analysis_machine_ngram}
\end{figure*}

%% file: sections/05-combine.tex
\section{Improving Neural LMs with the \methodname{}}
\label{sec:combine}

The results in \S\ref{sec:analysis} motivate us to combine neural LMs and \methodname{} (\S\ref{sec:method}) to yield better language models.
In this section, we will show strong experimental results of the combined model.
In \S\ref{sec:analysis} we found that the \methodname{} estimate has higher agreement with human-written text when it is \textit{sparse}.
Therefore, we use two separate interpolation hyperparameters: $\lambda_1$ for sparse and $\lambda_2$ for non-sparse \methodname{} estimates.
These hyperparameters are tuned on the validation set to minimize the perplexity of the combined model.

\subsection{Experimental setup}
\label{subsec:exp-setup}

\tightparagraph{Evaluation and metric.}
We measure the perplexity of each model on Pile's validation and test sets, as well as the relative improvement of perplexity between models.
To show generalization, we also evaluate on time-shifted data (i.e., data created after the cutoff date of the \methodname{} training data).
The relative improvement of model $M$ against model $M_o$ is defined as $(1 - \frac{\text{PPL}(M) - 1}{\text{PPL}(M_o) - 1}) \times 100\%$, which is the percentage of perplexity gap closed towards perfect language modeling (i.e., PPL = 1).
Additional details on evaluation data processing in \S\ref{subapp:exp-setup}.

\tightparagraph{Reference data.}
To reduce confusion, in this section we will use \textit{reference data} to refer to the training data of the \methodname{}.
In addition to Pile's training set that we used in the previous analyses (\S\ref{sec:analysis}), we also consider RedPajama \citep{together2023redpajama} as reference data.
The decontaminated Pile-train and Redpajama have 360 billion and 1.4 trillion tokens, respectively, summing up to 1.8 trillion tokens (based on the Llama-2 tokenizer).
We later perform ablations on varying sizes and domains of the reference data.

\tightparagraph{Neural LMs.}
We use a range of large, competitive neural LMs, both as baselines and as models to interpolate with the \methodname{}.
In total, 14 models are considered: GPT-2 117M/345M/774M/1.6B \citep{radford2019language}, GPT-Neo 125M/1.3B/2.7B, GPT-J-6B, Llama-2 7B/13B/70B, and SILO PD/PDSW/PDSWBY \citep{silo}.
See \S\ref{subapp:exp-setup} for additional details about these models and their training data.

\tightparagraph{Tokenizers.}
Among these models, GPT-2, GPT-Neo and GPT-J share the same tokenizer, but Llama-2 and SILO use different tokenizers.
We therefore built three versions of the \infinigram{} index on Pile-train and RedPajama, one for each tokenizer.
Due to the tokenizer variation, the perplexity of GPT-2, GPT-Neo and GPT-J are comparable to each other, but perplexity of Llama-2 and SILO are not comparable to them nor to each other.

\subsection{Results}
\label{subsec:exp-results}

\input{floats/result_ppl}

Experimental results with GPT-2, GPT-Neo/J, and Llama-2 on Pile's evaluation sets are shown in \autoref{tab:results_ppl}.
Results on time-shifted data can be found in \S\ref{subapp:timeshift}, and we show the impact of the size and domain of reference data in \S\ref{subapp:ablations}.

Interpolating with \methodname{} greatly and consistently improves the perplexity of neural LMs.
The magnitude of improvement trends smaller as the neural LM size grows within the same family, while the largest models can still benefit a lot from our method (e.g., Pile-train alone improves Llama-2 70B by 12\%).

However, this trend does not hold across different families of LMs.
For example, \methodname{} can improve GPT-2 1.6B by 34\%, but only improves a smaller model, GPT-Neo 1.3B, by 16\%.
This may be because GPT-Neo/J models are trained precisely on Pile, while GPT-2 models are not.
\methodname{} works better when the reference data distribution differs from, or complements, the pretraining data distribution, which emphasizes the importance of data diversity.
Meanwhile, the fact that \methodname{} also improves neural LMs already pretrained on its reference data shows that there is consistent advantage in introducing \methodname{}.

On the choice of \methodname{} reference data, the union of Pile-train and RedPajama yields larger improvements on the Llama-2 models than Pile-train alone.
The combination of Llama-2 13B and \methodname{} with Pile-train + RedPajama outperforms Llama-2 70B, and interpolating with \methodname{} pushes the perplexity of Llama-2 70B below 4.0.

\input{floats/result_silo}

When the neural LM is SILO (which is trained on permissive-licensed data only and thus has less training data), adding the \methodname{} component is more helpful when SILO is trained on more restrictive data (i.e., PD $>$ PDSW $>$ PDSWBY).
The usage of \methodname{} can be precisely traced back to the contributing document(s) in the reference data, which is in-line with the philosophy of SILO: to allow crediting the source data when using them for language modeling.
When compared to the existing retrieval-augmentation methods used by SILO, i.e., $k$NN-LM and RIC-LM, \methodname{} yields better improvement in perplexity.
Therefore, \methodname{} can serve as a better alternative as the retrieval-augmentation method for SILO.

\tightparagraph{A note on text generation.}
While \methodname{} can be interpolated with neural LMs and greatly improve their perplexity, our preliminary experiments show that such method might not be helpful, and even harmful, to open-ended text generation tasks.
During generation, \methodname{} can make odd mistakes (e.g., predicting totally irrelevant tokens) which makes the model to digress.
Thus this combined model is \textit{not} ready to replace neural LMs.
Additional investigation is required to make \methodname{} best contribute to text generation. 

%% file: floats/result_ppl.tex
\begin{table*}[!t]
\footnotesize
\centering
\resizebox{0.85 \linewidth}{!}{%
\begin{tabular}{lrl rrl rrl}
\toprule
    \multirow{2}{*}{\textbf{Neural LM}} &
    \multirow{2}{*}{\textbf{Size}} &
    \multirow{2}{*}{\textbf{Reference Data}} &
    \multicolumn{3}{c}{\textbf{Validation}} &
    \multicolumn{3}{c}{\textbf{Test}} \\
    \cmidrule(lr){4-6} \cmidrule(lr){7-9}
    & & & \textbf{Neural} & \multicolumn{2}{c}{\textbf{+ \methodname{}}}
    & \textbf{Neural} & \multicolumn{2}{c}{\textbf{+ \methodname{}}} \\
\midrule
GPT-2   & 117M & Pile-train         & 22.82 & \textbf{13.71} & \rel{42}  & 22.86 & \textbf{13.58} & \rel{42} \\
GPT-2   & 345M & Pile-train         & 16.45 & \textbf{11.22} & \rel{34}  & 16.69 & \textbf{11.18} & \rel{35} \\
GPT-2   & 774M & Pile-train         & 15.35 & \textbf{10.39} & \rel{35}  & 15.40 & \textbf{10.33} & \rel{35} \\
GPT-2   & 1.6B & Pile-train         & 14.42 & \textbf{ 9.93} & \rel{33}  & 14.61 & \textbf{ 9.93} & \rel{34} \\
\midrule
GPT-Neo & 125M & Pile-train         & 13.50 & \textbf{10.76} & \rel{22}  & 14.08 & \textbf{10.79} & \rel{25} \\
GPT-Neo & 1.3B & Pile-train         &  8.29 & \textbf{ 7.31} & \rel{13}  &  8.61 & \textbf{ 7.36} & \rel{16} \\
GPT-Neo & 2.7B & Pile-train         &  7.46 & \textbf{ 6.69} & \rel{12}  &  7.77 & \textbf{ 6.76} & \rel{15} \\
GPT-J   & 6.7B & Pile-train         &  6.25 & \textbf{ 5.75} & \rel{10}  &  6.51 & \textbf{ 5.85} & \rel{12} \\
\midrule
Llama-2 &  7B & Pile-train          &  5.69 & \textbf{ 5.05} & \rel{14}  &  5.83 & \textbf{ 5.06} & \rel{16} \\
Llama-2 & 13B & Pile-train          &  5.30 & \textbf{ 4.75} & \rel{13}  &  5.43 & \textbf{ 4.76} & \rel{15} \\
Llama-2 & 70B & Pile-train          &  4.59 & \textbf{ 4.21} & \rel{11}  &  4.65 & \textbf{ 4.20} & \rel{12} \\
\midrule
Llama-2 &  7B & Pile-train + RedPajama    &  5.69 & \textbf{ 4.66} & \rel{22}  &  5.83 & \textbf{4.66 } & \rel{24} \\
Llama-2 & 13B & Pile-train + RedPajama    &  5.30 & \textbf{ 4.41} & \rel{21}  &  5.43 & \textbf{4.42 } & \rel{23} \\
Llama-2 & 70B & Pile-train + RedPajama    &  4.59 & \textbf{ 3.96} & \rel{18}  &  4.65 & \textbf{3.95 } & \rel{19} \\
\bottomrule
\end{tabular}
}%
\vspace{-5pt}
\caption{
    Perplexity (lower is better) on Pile's validation and test sets.
    Numbers in parentheses are the relative perplexity improvement.
    The first eight rows share the same tokenizer, and the last six rows share the same tokenizer.
}
\vspace{-12pt}
\label{tab:results_ppl}
\end{table*}

%% file: floats/result_silo.tex
\begin{table*}[!t]
\myfontsize
\centering
\setlength{\tabcolsep}{1pt}
\begin{tabular}
    {p{2cm} R{1cm}R{1cm}lrr R{1cm}R{1cm}lrr}
\toprule
    \multirow{2}{*}{\textbf{Neural LM}} &
    \multicolumn{5}{c}{Validation} &
    \multicolumn{5}{c}{Test} \\
    \cmidrule(lr){2-6} \cmidrule(lr){7-11}
    & \textbf{Neural} & \multicolumn{2}{c}{\textbf{+ \methodname{}}}
    & + \knnlm{}$^\dagger$ & + RIC-LM$^\dagger$
    & \textbf{Neural} & \multicolumn{2}{c}{\textbf{+ \methodname{}}}
    & + \knnlm{}$^\dagger$ & + RIC-LM$^\dagger$ \\
\midrule
    \multicolumn{7}{l}{\textbf{\em Eval data: Wikipedia}} \\
    Silo PD          & \OD{26.60} &  \OD{\textbf{15.30}} & \ \ODRel{43} & \OD{20.62} & \OD{27.91} &  \OD{28.42} &  \OD{\textbf{14.44}} & \ODRel{51} & \OD{--} & \OD{--} \\
    Silo PDSW        & \OD{18.93} &  \OD{\textbf{12.36}} & \ \ODRel{36} & \OD{14.10} & \OD{18.90} &  \OD{20.02} &  \OD{\textbf{11.84}} & \ODRel{43} & \OD{14.5} & \OD{19.4} \\
    Silo PDSWBY      & \ID{10.66} &  \ID{\textbf{ 8.77}} & \ \IDRel{19} & \ID{10.14} & \ID{10.87} &  \ID{10.76} &  \ID{\textbf{ 8.41}} & \IDRel{24} & \ID{--} & \ID{--} \\
    Pythia      & \ID{9.00} &  \ID{\textbf{}} & \ID{--} & \ID{8.50} & \ID{8.84} &  \ID{9.1} &  \ID{\textbf{}} & \ID{--} & \ID{--} & \ID{--} \\
\midrule
    \multicolumn{7}{l}{\textbf{\em Eval data: Enron Emails}} \\
    Silo PD          & \OD{19.56} &  \OD{\textbf{ 6.31}} & \ \ODRel{70} & \OD{8.56} & \OD{15.45} &  \OD{15.71} &  \OD{\textbf{ 4.85}} & \ODRel{73} & \OD{--} & \OD{--} \\
    Silo PDSW        & \OD{14.66} &  \OD{\textbf{ 5.58}} & \ \ODRel{65} & \OD{6.70} & \OD{10.80} &  \OD{11.23} &  \OD{\textbf{ 4.35}} & \ODRel{66} & \OD{5.9} & \OD{9.9} \\
    Silo PDSWBY      & \OD{14.67} &  \OD{\textbf{ 5.61}} & \ \ODRel{65} & \OD{7.24} & \OD{10.91} &  \OD{11.52} &  \OD{\textbf{ 4.44}} & \ODRel{66} & \OD{--} & \OD{--} \\
    Pythia      & \ID{7.577} &  \ID{\textbf{}} & \ID{--} & \ID{4.99} & \ID{6.16} &  \ID{6.9} &  \ID{\textbf{}} & \ID{--} & \ID{--} & \ID{--} \\
\midrule
    \multicolumn{7}{l}{\textbf{\em Eval data: NIH ExPorters }} \\
    Silo PD          & \OD{27.46} &  \OD{\textbf{16.26}} & \ \ODRel{41} & \OD{19.27} & \OD{25.51} &  \OD{27.94} &  \OD{\textbf{16.00}} & \ODRel{44} & \OD{--} & \OD{--} \\
    Silo PDSW        & \SD{19.35} &  \SD{\textbf{12.70}} & \ \SDRel{35} & \SD{14.95} & \SD{18.35} &  \SD{19.12} &  \SD{\textbf{12.39}} & \SDRel{37} & \SD{15.0} & \SD{18.5} \\
    Silo PDSWBY      & \SD{15.01} &  \SD{\textbf{10.62}} & \ \SDRel{30} & \SD{12.33} & \SD{14.29} &  \SD{14.81} &  \SD{\textbf{10.33}} & \SDRel{32} & \SD{--} & \SD{--} \\
    Pythia      & \ID{11.20} &  \ID{\textbf{}} & \ID{--} & \ID{11.20} & \ID{10.83} &  \ID{11.1} &  \ID{\textbf{}} & \ID{--} & \ID{--} & \ID{--} \\
\bottomrule
\end{tabular}
\caption{
    Perplexity (the lower the better) on the validation and the test datasets of the Wikipedia, Enron Emails, and NIH ExPorters of the Pile.
    All neural models are 1.3B models, and the reference data is always the Pile.
    \textcolor{bluex!25}{$\blacksquare$} indicates in-domain; \textcolor{orange!25}{$\blacksquare$}
    indicates out-of-domain; \textcolor{yellow!25}{$\blacksquare$} indicates out-of-domain but has relevant data in-domain, all with respect to the training data of the neural LM.
    $\dagger$: Results retrived from \citet{silo}, which use much smaller reference data: 45-million to 1.2-billion tokens, compared to our 360-billion tokens.
}
\label{tab:results_silo}
\end{table*}

%% file: sections/06-related.tex
\section{Related Work}
\label{sec:related}

We discuss closely related work here.
See \S\ref{app:discuss} for extended discussion, and \autoref{tab:related} for comparison with other \ngram{} models and nonparametric language models.

\tightparagraph{\ngrambold{} language models.}
\ngram{} has been one of the most classical language modeling methods since the inception of natural language processing \citep{Jurafsky2000SpeechAL}.
People have been pushing the limits of \ngram{} LMs by scaling up its training data.
To date, the largest \ngram{} table \citep{Brants2007LargeLM} counts 5-grams in a corpus of 2 trillion tokens.

While \ngram{} LMs are currently largely surpassed by neural LMs, there has been recent work that revisit \ngram{}s and \ngram{} LMs.
\citet{Mikolov2012ContextDR} finds that interpolating with a 5-gram Kneser-Ney model improves the perplexity of RNN models, whereas \citet{khandelwal2020generalization} finds that interpolating \ngram{} models with Transformers does not improve perplexity substantially.
\citet{li-etal-2022-residual} finds that the \ngram{} model is as competitive as a small neural LM, and training a neural model to be complementary to the \ngram{} model and using both at inference time outperforms the neural-only LM.
However, both use limited reference data (101M tokens) and compare with small neural LMs (117--250M parameters).
Some prior work has found value in scaling up the \ngram{} training data \citep{Allamanis2013MiningSC}.

Our work scales up the training data of \ngram{} LMs to trillions of tokens.
With the addition of scaling up the value of $n$ in the \ngram{}, our model can significantly improve state-of-the-art neural models as large as 70B.

\tightparagraph{Unbounded \ngrambold{}s, suffix arrays, suffix trees.}
Previous work has explored using suffix-based data structures to enable $n$-gram queries with unbounded $n$, with limited scale of the training data.
\citet{Stehouwer2010UsingSA} proposes to use suffix arrays for \methodname{}, and yet their formulation does \textit{not} yield proper probability distributions and, consequently, a language model.
\citet{Kennington2012SuffixTA} proposes to use suffix trees for the same purpose, and yet the storage overhead of suffix trees is very high such that it hinders scaling, which may be mitigated with highly intricate compression techniques \citep{Shareghi2015CompactEA}.
Among the three aforementioned papers, only the third evaluates on the general language modeling task, and the perplexity numbers are too high to be practically useful.
Our training data is 500x larger than the largest one in these previous work.

\tightparagraph{Nonparametric language models.}
Nonparametric LMs refer to LMs whose complexity is not bounded a priori, because the complexity can change according to the reference data \citep{khandelwal2020generalization,borgeaud2022improving,retrieval-lm-tutorial}.
The \methodname{} LM is one instance of nonparametric LMs, and its simplicity makes it possible to significantly scale the reference data with modest resources (\S\ref{sec:infra}).
To the best of our knowledge, our \methodname{} LM is the largest in both the size of the reference data (5 trillion tokens) and the size of the base neural LM (70B).

%% file: sections/07-concl.tex
\section{Conclusion}
\label{sec:concl}

In this paper, we modernized the classical \ngram{} language model by scaling it up to a trillion tokens and extending to unbounded $n$.
We presented the \infinigram{} engine that performs efficient training and inference under this extreme setup.
We also proposed the \methodname{} language model, powered by the \infinigram{} engine, and showed that it can offer novel insights into human-written and machine-generated text and can improve existing neural language models.

%% file: sections/09-ack.tex
\section*{Acknowledgments}

We would like to thank Zexuan Zhong, Mike Lewis, Yanai Elazar, Will Merrill, Tim Dettmers, Ximing Lu, Alisa Liu, Weijia Shi, Xiaochuang Han, members of the H2lab, and Ziqi Ma for their invaluable feedback.

%% file: sections/10-app.tex
\section{Additional Details on the \Infinigram{} Engine}
\label{app:method}

\input{floats/sa2}

\subsection{What is the size of the \ngram{} count table implied by an \infinigram{} index?}
\label{subapp:table-size}

The size of an \ngram{} LM is often measured by the number of unique \ngram{}s indexed, each associated with its count in the dataset.
This size is easy to obtain when indexing is done as the classical \ngram{} count tables, but non-trivial to compute for \infinigram{}.

If we consider all possible values of $n$ in the \ngram{}, then a dataset with $N$ tokens would contain about $\frac{1}{2} N^2$ \ngram{}s, and since most of these \ngram{}s are long and thus very likely to be distinct, there would be about $\frac{1}{2} N^2$ unique \ngram{}s.
Since we index $N = $ 5 trillion tokens, the size of the \ngram{} count table implied by our \infinigram{} index would be approximately $1.2 \times 10^{25}$, or equivalently, 20 \textit{mol} (using $N_A = 6.02 \times 10^{23} / mol$).

However, the document separator is meaningless and should not be part of the \ngram{}s, and we should probably only count \ngram{}s within the document boundaries.
There are $N = 5 \times 10^{12}$ tokens and $D = 6 \times 10^9$ documents, and on average each documents have 857 tokens.
Using the Cauchy-Schwarz inequality, we have the total number of \ngram{}s as
$$
\sum_d \frac{1}{2} N_d^2 \ge \frac{1}{2} D \cdot (\frac{N}{D})^2 = \frac{N^2}{2 D} = 2 \times 10^{15}
$$
Therefore, there are at least 2 quadrillion unique \ngram{}s in the count table implied by \infinigram{}.

\subsection{Additional details on the \infinigram{} index}
\label{subapp:sa}

\tightparagraph{\Infinigram{} indexes are additive and subtractive.}
If we have two or more indexes built on disjoint datasets (with the same tokenizer), we can easily combine them into a single index by adding up the \ngram{} counts from each index.
This feature is useful in the sharding technique that we discuss in \S\ref{subapp:indexing} and \S\ref{subapp:inference}.
Similarly, if we have built indexes on a big dataset and a subset of it, we can easily obtain an index of their difference by taking the difference of \ngram{} counts.
Compared to having a single index, both operations incur some additional inference operations (which can be parallelized to mitigate latency overhead), but they would spare us from re-indexing the union or difference sets from scratch.

\tightparagraph{Document offsets and metadata.}
To enable efficient document retrieval, the \infinigram{} index stores additional data about documents.
A \textit{document offset} file stores the byte offset of each document in the tokenized dataset, and its format is similar to the suffix array.
A \textit{document metadata} file stores a comma-separated string for each document that contains its metadata (e.g., document ID, source, URL), and a \textit{document metadata offset} file stores the byte offset of each document's metadata in the document metadata file.
All the above files are negligible in size compared to the suffix array, because there are far less documents than the total number of tokens.

\subsection{Additional details on building the suffix array}
\label{subapp:indexing}

\tightparagraph{Sharding.}
Building the suffix array requires heavy random access to the byte array, and thus the entire byte array must be kept in RAM so that the building time is reasonable.
However, the byte array may be too large to fit into RAM.
In such cases, we shard the byte array into multiple shards, and build a suffix array for each shard.
Sharding would induce additional inference latency, which we discuss and mitigate below (\S\ref{subapp:inference}).

\subsection{Additional details on inference with the \infinigram{} index}
\label{subapp:inference}

Both the first and last occurrence positions can be found with binary search, with time complexity $O(n \cdot \log{N})$ and $O(\log{N})$ random array accesses.
The two binary searches can be parallelized, reducing the latency by roughly 2x.
The impact of query length $n$ is negligible, because computers usually fetch memory in pages of 4K bytes, and string comparison is much faster than page fetching.
Therefore, when we analyze time complexity below, we refer to the number of random array accesses.

\tightparagraph{Finding occurrence positions and documents.}
\ngram{} counting with suffix arrays has a by-product: we also get to know all positions where the \ngram{} appears in the training data, for free.
This position information is implicitly contained in the suffix array segment we obtained during counting, and to retrieve the original documents where the \ngram{} appears, all we need to do is to follow each pointer within this segment back into the tokenized dataset, and find the starting and ending position of the enclosing document by performing a binary search on the \textit{document offset} index.
(Note that if we don't have the \textit{document offset} index, the latency of document search cannot be bounded because we would need to expand the pointer in both directions in the tokenized dataset until hitting the document separator.
In practice, we see documents as large as 20M tokens.)

\tightparagraph{Impact of sharding.}
When the suffix arrays are built on sharded byte arrays, we can simply perform counting on each individual shard and accumulate the counts across all shards.
The latency is proportional to the number of shards: time complexity would become $O(S \cdot \log{N})$.
The processing of different shards can be parallelized, reducing the time complexity back to $O(\log{N})$.

\tightparagraph{Speeding up $n$-gram computation by re-using previous search results.}
On the suffix array, the segment for $x_1 ... x_n$ must be a sub-segment of that for $x_1 ... x_{n-1}$.
Therefore, when computing the $n$-gram probability $P_n(x_n \mid x_1 ... x_{n-1})$, we can first count $x_1 ... x_{n-1}$, and then when counting $x_1 ... x_n$, we only need to search for the first and last occurrence positions within the segment of $x_1 ... x_n$, which reduces the latency by at most 2x.

\tightparagraph{On-disk search.}
The byte array and suffix array may be too large to fit into RAM, so in practice, we keep them on disk and read them as memory-mapped files.
However, this creates a significant latency as the binary search requires random access to the byte array and suffix array.
To mitigate this, we implemented a memory \textbf{pre-fetching} method that informs the system of the array offsets we will likely be reading in the near future.
Pre-fetching reduces average latency by roughly 5x.

\tightparagraph{Speeding up \methodname{} computation.}
To compute the \methodname{} probability, we need to count the occurrence of each suffix $x_{l-n+1} ... x_l$ up to the maximum $n$ so that the suffix still meets the sufficient appearance requirement (we denote this maximum $n$ as $L$).
This means $O(L)$ counting operations, and the time complexity for each \methodname{} computation is $O(L \cdot \log{N})$.
However, a simple binary-lifting + binary-search algorithm for searching $L$ can reduce the number of counting operations to $O(\log{L})$, and thus the time complexity for each \methodname{} computation becomes $O(\log{L} \cdot \log{N})$.

\tightparagraph{Speeding up dense \methodname{} computation.}
During evaluation, we need to compute the \methodname{} probability of each token in the test document.
We can save computation by observing that the effective $n$ for one token is at most one token longer than that for the previous token.
This brings the amortized time complexity for evaluating each token down to $O(\log{N})$.

\subsection{Supported query types and latency benchmarking}
\label{subapp:query_types}

\Infinigram{} supports the following types of \ngram{}/\methodname{} queries:
\begin{enumerate}
\item Counting an \ngram{} (\textsc{Count});
\item Computing a token probability from \ngram{} LM (with given $n$, no backoff) (\textsc{NgramProb});
\item Computing the full next-token distribution from \ngram{} LM (\textsc{NgramDist});
\item Computing a token probability from \methodname{} LM (\textsc{InfgramProb});
\item Computing the full next-token distribution from \methodname{} LM (\textsc{InfgramDist});
\item Returning documents containing an \ngram{}, or a CNF logical expression of \ngram{} terms, connected with AND's and/or OR's (e.g., \texttt{(natural language processing OR artificial intelligence) AND (deep learning OR machine learning)}) (\textsc{SearchDoc}).
\end{enumerate}

\input{floats/benchmarking}

We benchmark the latency of \infinigram{} on different types of $n$-gram and \methodname{} queries, and show results in \autoref{tab:benchmarking}.
During inference, the training data and the suffix array are stored on an SSD.
For each type of query, the benchmarking is conducted on 1,000 tokens randomly and independently sampled from Pile's validation data (except for the task ``computing a token probability from \methodname{} LM on consecutive tokens'', where we sampled 10 documents and processed 1000 consecutive tokens in each document).

All types of queries demonstrate sub-second latency on the trillion-token training data.
Computing a token probability from the \methodname{} with RedPajama takes merely 135 milliseconds.
Furthermore, our implementation supports counting the occurrence of an $n$-gram with arbitrarily large $n$, with roughly constant latency at 20 milliseconds (we experimentally validated up to $n = 1000$).
Decoding requires computing the full next-token distribution and is thus slightly slower: 39 milliseconds per token with \ngram{} LMs and 180 milliseconds per token with \methodname{}.

\section{Decontamination of Reference Data}
\label{app:decontamination}

To properly evaluate the effectiveness of \methodname{} LM on Pile's evaluation sets, we performed data decontamination on the Pile's training set and RedPajama before using them as reference data for the \methodname{} LM.
We run the Big Friendly Filter (BFF)\footnote{\url{https://github.com/allenai/bff}} \citep{bff} on Pile's training set and RedPajama, filtering out documents with too much $n$-gram overlap with Pile's evaluation sets.
\autoref{tab:bff_stats} reports the statistics of decontamination.

When using BFF, we always remove whole documents, instead of by paragraphs.
Following the default settings, we consider $n$-grams where $n = 13$, and discard the document if at least 80\% of its $n$-grams are present in the evaluation set.
For Pile's training set, we lowercase all documents to capture more potential contaminations.

\input{floats/bff_stats}

\section{Additional Analysis}
\label{app:analysis}

\input{floats/analysis_human_ngram_more}
\input{floats/analysis_human_neural_ngram_more}
\input{floats/analysis_machine_ngram_more}

\autoref{fig:analysis_human_ngram_more} is an extension to the middle plot of \autoref{fig:analysis_human_ngram}, and shows a more fine-grained analysis of the token-wise agreement between human-written text and \methodname{}.
\autoref{fig:analysis_human_neural_ngram_more} extends \autoref{fig:analysis_human_neural_ngram} with results on Llama-2-13b/7b.
\autoref{fig:analysis_machine_ngram_more} extends \autoref{fig:analysis_machine_ngram} with results on GPT-Neo models.

\section{Additional Experiments on Improving Neural LMs with \methodname{}}
\label{app:ppl_results}

\subsection{Additional details on experimental setup}
\label{subapp:exp-setup}

\tightparagraph{Evaluation data processing.}
We split each document in the evaluation data into batches with a maximum sequence length of 1024 and a sliding window of 512, a setup that is standard in prior language modeling literature \citep{baevski2018adaptive,khandelwal2020generalization}.

\tightparagraph{Neural LMs.}
Below are additional details about the families of neural LMs we use.
\begin{itemize}[itemsep=2pt, leftmargin=8pt, topsep=1pt]
\item \textbf{GPT-2}~\citep{radford2019language}, one of the earliest autoregressive language models whose sizes range from 117M, 345M, and 774M to 1.6B. Their training data is a diverse set of web text, although is not public.
\item \textbf{GPT-Neo}~\citep{gao2020pile} and \textbf{GPT-J}~\citep{gpt-j}, language models trained on the Pile whose sizes vary from 125M, 1.3B, and 2.7B to 6.7B.
\item \textbf{Llama-2}~\citep{touvron2023llamatwo}, a subsequent version of LLaMA~\citep{touvron2023llama} trained on two trillion tokens and has sizes of 7B, 13B, and 70B. Llama-2 is one of the most competitive language models whose weights are available at the time of writing the paper. The training data of Llama-2 is unknown, although the precedent version is trained on a large corpus of Common Crawls, Wikipedia and code, which is replicated by RedPajama~\citep{together2023redpajama}.
\item \textbf{SILO}~\citep{silo}, 1.3B language models trained on permissively licensed data only. The original paper showed that training on permissively licensed data leads to the challenge of {\em extreme domain generalization} because the training data is skewed to highly specific domains like code and government text. We use three different variants, PD, PDSW and PDSWBY, which are trained on different levels of permissivity, leading to varying levels of the domain generalization challenge.
\end{itemize}

\subsection{Evaluating on time-shifted data}
\label{subapp:timeshift}

\input{floats/result_timeshift}

To further show the effectiveness of \methodname{} and eliminate doubts that our performance gains might be due to insufficient decontamination, we evaluate on \textbf{time-shifted data}: documents that were created after the cutoff time of the \methodname{} reference data.
We use new Wikipedia articles created during April and August, 2023, which is after the cutoff time of both Pile and RedPajama.

\autoref{tab:result_timeshift} reports the perplexity of neural LM as well as the combined model.
On documents in four out of the five months, interpolating with \methodname{} improves the perplexity of the neural LM.
We find that this improvement can be further boosted by applying a Random Forest to decide an instance-wise interpolation hyperparameter, where the features of the Random Forest are the suffix lengths (1 up to the effective $n$) as well as the frequency of each suffix in the reference data.
When Random Forest is applied, the perplexity improvement ranges from 3\% -- 20\%.

\subsection{Ablations}
\label{subapp:ablations}

\input{floats/ablation_scale}

\tightparagraph{Effect of the size of reference data.}
\autoref{fig:ablation_scale} reports performance of the combined model wrt the size of reference data.
To create progressively smaller reference data, we repeatedly downsampled the full reference data by 2x (up to 256x, resulting in 9 sizes).
We see the improvement brought by \methodname{} widens as reference data size grows, and the relationship is roughly log-linear (except for the \texttt{NIH ExPorter} domain, where \methodname{} doesn't help when the reference data is too small).

\tightparagraph{Effect of the domain of reference data.}
\autoref{fig:ablation_scale} also compares performance of the combined model where the \methodname{} uses either the full reference data or only the in-domain reference data.
Using only the in-domain reference data is roughly as powerful as using the full reference data, which implies that almost all improvement we have achieved is thanks to in-domain data (which has been decontaminated).
This means it would not hurt to use the full reference data, especially when the test domain is unknown or an in-domain reference data is unavailable; however, having in-domain reference data is most helpful.

\section{Extended Discussion}
\label{app:discuss}

In \S\ref{sec:analysis} and \S\ref{sec:combine}, we showcased some very preliminary use cases of the \infinigram{} engine.
However, we believe that \infinigram{} can enable much broader investigations and applications, including but not limited to:

\tightparagraph{Understanding massive text corpora.}
Text corpora used for pretraining language models have become prohibitively large, and we have relatively limited understanding of their contents \citep{Elazar2023WhatsIM}.
\Infinigram{} can be a useful tool to quickly find out what \textit{is} in the corpus and what is \textit{not}, using \ngram{} lookup (the \textsc{Count} query).

\tightparagraph{Data curation.}
Data engineers often want to remove problematic content in corpora scraped from the Internet, such as toxicity, hate speech, and personal identifiable information (PII).
Using \infinigram{}'s \textsc{SearchDoc} query (which can be easily modified to return all documents), one can retrieve all documents containing an \ngram{} term (or a CNF expression with multiple \ngram{} terms) and remove them from the corpus.
Removal can even be done iteratively: \infinigram{} indexes are additive/subtractive, so we can obtain an index of the corpus after round one removal, by indexing the removed set and take the difference of the original index and the removal index.

\tightparagraph{Document retrieval.}
The scale of datastore is key to the effectiveness of retrieval-augmented LMs \citep{Asai2024ReliableAA}.
However, vector-based indexes are difficult to scale up due to compute limit, storage limit, and inference efficiency.
\Infinigram{}'s \textsc{SearchDoc} function can retrieve documents from datastores as large as the full pretraining corpora, and can potentially boost the performance of retrieval-augmented LMs.

\tightparagraph{Reducing hallucination in factual knowledge.}
Parametric-only models are prone to generating non-factual statements, which is widely known as the hallucination problem.
\Infinigram{} can potentially be used to mitigate hallucination by reading verbatim from the training data.
We have found evidence that the \methodname{} can greatly outperform Llama-2-70B on factual probing benchmarks such as LAMA \citep{Petroni2019LanguageMA}.

\tightparagraph{Detecting data contamination, memorization, and plagiarism.}
Test set contamination has become a major issue for language model evaluation.
\ngram{} lookup enables us to check if evaluation queries have sneaked into the training data of neural LMs.
It also opens up possibility to detect memorization in machine-generated text, or plagiarism in human-written text.

\tightparagraph{Preventing copyright infringement.}
Recently, generative AIs are facing numerous lawsuits for generating arguably copyrighted materials.
\Infinigram{} may be helpful for preventing copyright infringement, by diverting neural LMs to alternative (yet still plausible) generation paths when they are about to generate long \ngram{}s that appear in the training data, especially if they mostly appear in documents from copyrighted sources.

\tightparagraph{Measuring popularity of entity.}
\citet{Mallen2022WhenNT} showed that LMs have better memorization of facts about popular entities than less popular ones.
In that paper, heuristic metrics like Wikipedia pageviews are used as proxy to popularity.
A better proxy may be the number of appearances of the entity's name in a massive text corpus, which can be easily computed by \infinigram{}'s \textsc{Count} query.

\tightparagraph{Measuring novelty and creativity of text.}
Are neural LMs generating genuinely novel text, or are they simply copying from its pretraining data?
We need a metric for text novelty and creativity, and this metric can be potentially defined by the $n$-gram overlap between the generated document and the pretraining corpus.

\tightparagraph{Attribution.}
When using neural LMs to make predictions, people might want to know which training data most influenced the model's decision.
Using \ngram{} lookup with key phrases, we can trace back to related documents in the training data of neural LMs.

\tightparagraph{Non-parametric speculative decoding.}
Speculative decoding \citep{Chen2023AcceleratingLL} speeds up text generation by employing a fast and a slow decoder, where the fast decoder is a smaller model that does the autoregressive token generation, and the slow decoder checks the fast decoder's proposals by parallelizing the forward passes of multiple tokens.
Given the low latency of \infinigram{}, we can potentially use \methodname{} as the fast decoder, similar to \citet{He2023RESTRS}.

We welcome the community to collaboratively build toward the aforementioned directions, by leveraging our publicly released web interface and API endpoint.

\section{Extended Related Work}
\label{app:related}

\input{floats/related}

\tightparagraph{\ngrambold{}s beyond $n = 5$.}
\citet{Mochihashi2007TheIM} and \citet{Wood2009ASM} consider infinitely-ordered Markov models, to which the \methodname{} LM is a special case.
Aside from the \methodname{} LM, some other work has found value in scaling the value of $n$ in \ngram{}s.
For example, KiloGrams \citep{Raff2019KiloGramsVL} uses 1024-grams as features for malware detection.

\tightparagraph{Other data structures for text indexing.}
Beside suffix arrays and suffix trees, other data structures have been used to index text corpora to satisfy different trade-offs.
Koala \citep{Vu2023KoalaAI} builds a compressed suffix array to analyze overlap between text corpora.
The ROOTS Search Tool \citep{Piktus2023TheRS} builds a BM25 index on the ROOTS corpus, and supports document searching via both exact match and fuzzy match of \ngram{}s.
Data Portraits \citep{Marone2023DataPR} proposes a lightweight index based on Bloom Filter, and is tailored for probabilistic membership inference (exact match of \ngram{}s of 50 characters, where $n \approx 8$) against the Pile and Stacks.
ElasticSearch is a proprietary search engine based on the Lucene index, and it has been used by \citet{Dodge2021DocumentingLW} to search documents in C4, and also by \citet{Elazar2023WhatsIM} to count \ngram{}s and list most frequent \ngram{}s in various corpora up to 480B tokens.

\tightparagraph{Nonparametric language models.}
A nonparametric LM refers to the LM whose complexity is not bounded as a priori, because the complexity can grow or update according to the data given at inference time.
Prior work is broadly divided into two categories:
a {\em token} retrieval approach that represents each token as one vector and uses a nonparametric prediction function \citep{khandelwal2020generalization, zhong2022training, lan2023copy, min2022nonparametric, silo, shi2023replug}, 
and a {\em chunk} retrieval approach that represents each chunk of text as a vector and incorporates nearest chunks to the neural language model \citep{guu2020retrieval, izacard2022few, borgeaud2022improving}. 
Scaling the reference data in nonparametric LMs is very expensive as it requires storing a vector for every unit (either token or chunk).
To the best of our knowledge, prior work with the largest reference data is \retro\ \citep{borgeaud2022improving}, which uses the 7B-parameter LM and the reference data consisting of 1.8 trillion tokens. It stores and searches over 28 billion vectors, estimated to consume 432TB of disk space.\footnote{This is in part because \retro\ does not use any approximation in \knn\ search. Even if \retro\ used approximate search as \citet{khandelwal2020generalization} did, it would still use 10TB. Moreover, there is no open-sourced software that easily supports fast \knn\ search over tens of billions of vectors.}
(Detailed comparisons in \autoref{tab:related}.)

\section{Example Queries and Web Interface}

Figures \ref{fig:queries_1} to \ref{fig:queries_6} show one example for each of the six query types supported by \infinigram{}.
We query the Dolma corpus in these examples.
Screenshots are taken from our web interface.

\input{floats/queries/1}
\input{floats/queries/2}
\input{floats/queries/3}
\input{floats/queries/4}
\input{floats/queries/5}
\input{floats/queries/6}

%% file: floats/sa2.tex
\begin{figure*}[!b]
\centering
\includegraphics[width=\linewidth]{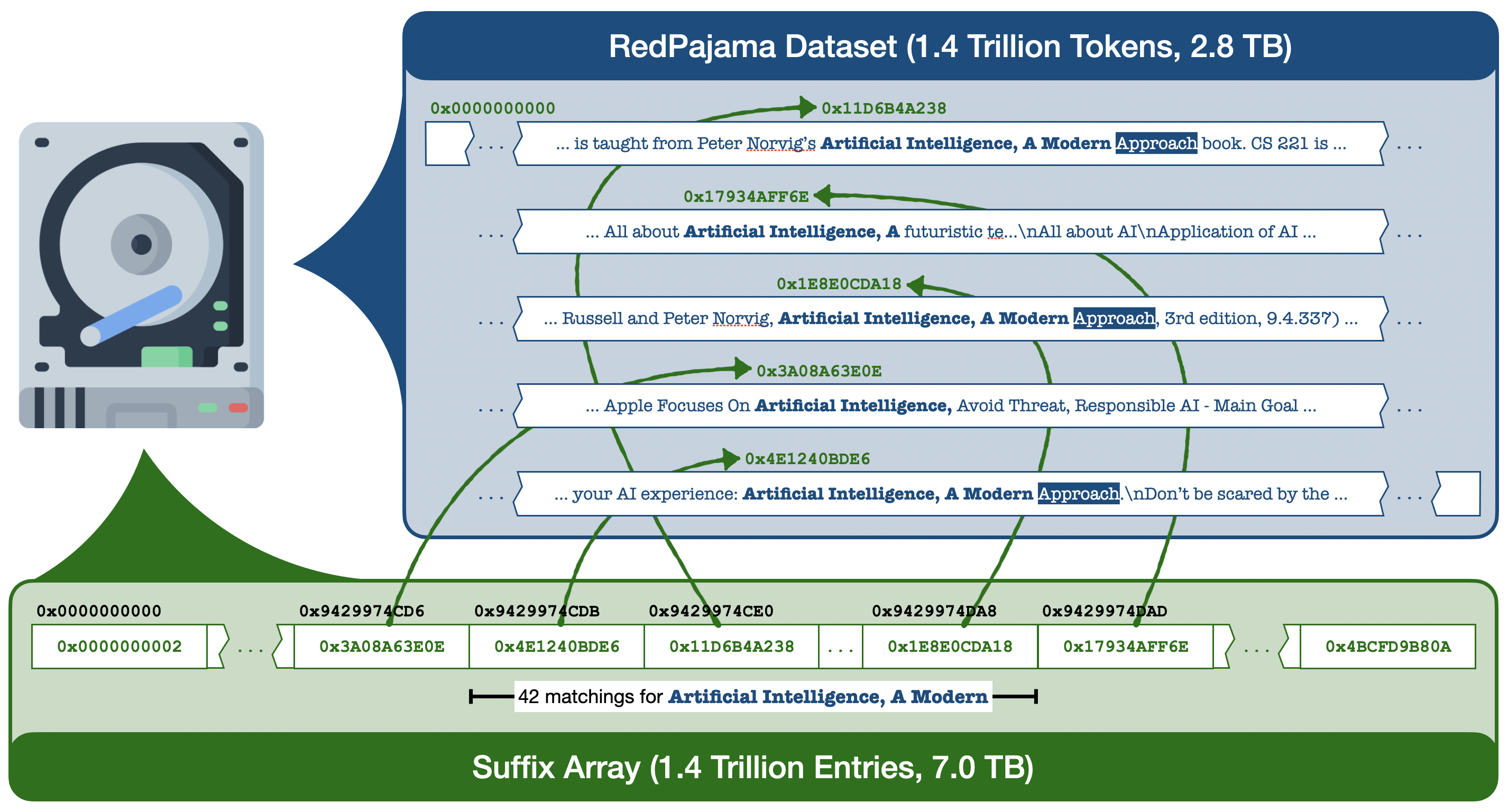}
\caption{
    \ngram{}/\methodname{} queries on a training data are supported by an associated \textit{suffix array}.
    Both the training data and the suffix array are stored on-disk as regular files.
    Contents on the white strips are file data, and addresses above the strips are byte offsets.
    Querying for a particular $n$-gram returns a consecutive segment of the suffix array, where each element is a pointer into the training data where the $n$-gram appears.
    E.g., in the trillion-token training data, \texttt{Artificial Intelligence, A Modern} appears 42 times, and in all cases the following token is \texttt{Approach}. 
}
\label{fig:sa2}
\end{figure*}

%% file: floats/benchmarking.tex
\begin{table*}[!t]
\centering
\setlength{\tabcolsep}{3pt}
\resizebox{\linewidth}{!}{%

\begin{tabular}{lccl}
\toprule
\hspace{8em} \textbf{Reference Data $(\rightarrow)$} & \textbf{Pile-train} & \textbf{RPJ} & \textbf{Time Complexity} \\
& $N = $ 0.36T & $N = $ 1.4T & (measured by number \\
\textbf{Query Type $(\downarrow)$} & $S = 2$ & $S = 8$ & of random disk accesses) \\
\midrule
1. Counting an $n$-gram & & & $O(\log N)$ \\
\hspace{1em} ... ($n = 1$) & 7 ms & 9 ms \\
\hspace{1em} ... ($n = 2$) & 13 ms & 20 ms \\
\hspace{1em} ... ($n = 5$) & 14 ms & 19 ms \\
\hspace{1em} ... ($n = 10$) & 13 ms & 18 ms \\
\hspace{1em} ... ($n = 100$) & 13 ms & 19 ms \\
\hspace{1em} ... ($n = 1000$) & 14 ms & 19 ms \\
2. Computing a token probability from $n$-gram LM ($n = 5$) & 19 ms & 30 ms & $O(\log N)$ \\
3. Computing full next-token distribution from $n$-gram LM (n = 5) & 31 ms & 39 ms & $O(V \cdot \log N)$ \\
4. Computing a token probability from \methodname{} LM & 90 ms & 135 ms & $O(\log L \cdot \log N)$ \\
\hspace{1em} ... on consecutive tokens & 12 ms & 20 ms & $O(\log N)$ \\
5. Computing full next-token distribution from \methodname{} LM & 88 ms & 180 ms & $O((\log L + V) \cdot \log N)$ \\
\bottomrule
\end{tabular}
}%
\caption{
    Inference-time latency of \infinigram{} on different types of queries.
    Average latency per query is reported.
    Benchmarked with inference engine written in C++ (with parallelized shard processing) and running on a single, 8-core CPU node.
    Notations for time complexity:
    $N = $ number of tokens in the reference data;
    $S = $ number of shards for the suffix array;
    $L = $ number of tokens in the query document;
    $V = $ vocabulary size.
}
\label{tab:benchmarking}
\end{table*}

%% file: floats/bff_stats.tex
\begin{table*}[!h]
\begin{minipage}{0.48 \textwidth}
\resizebox{\linewidth}{!}{%
\begin{tabular}{lrrl}
\toprule
\multicolumn{4}{c}{\textsc{RedPajama}} \\
\midrule
\textbf{Subset} & \textbf{Total docs} & \textbf{Filtered docs} & \textbf{Ratio filtered} \\
\midrule
arxiv & 1558306 & 213 & 0.01\% \\
book & 205744 & 711 & 0.3\% \\
c4 & 364868892 & 53195 & 0.01\% \\
common\_crawl & 476276019 & 0 & 0\% \\
github & 28793312 & 614259 & 2\% \\
stackexchange & 29825086 & 40086 & 0.01\% \\
wikipedia & 29834171 & 21973 & 0.07\% \\
\midrule
\textbf{Total} & \textbf{931361530} & \textbf{730437} & \textbf{0.08\%} \\
\bottomrule
\end{tabular}
}%
\end{minipage}
\hfill
\begin{minipage}{0.51 \textwidth}
\resizebox{\linewidth}{!}{%
\begin{tabular}{lrrl}
\toprule
\multicolumn{4}{c}{\textsc{Pile (train)}} \\
\midrule
\textbf{Subset} & \textbf{Total docs} & \textbf{Filtered docs} & \textbf{Ratio filtered} \\
\midrule
Arxiv & 2377741 & 1089 &  \\
BookCorpus2 & 25355 & 6 &  \\
Books3 & 277655 & 99 &  \\
DM Mathematics & 1918535 & 0 & 0\% \\
Enron Emails & 926132 & 18236 & 2\% \\
EuroParl & 131723 & 21 &  \\
FreeLaw & 5069088 & 11821 & 0.2\% \\
Github & 18044218 & 961726 & 5.3\% \\
Gutenberg (PG-19) & 66981 & 70 & 0.1\% \\
HackerNews & 1571968 & 14 &  \\
NIH ExPorter & 1777926 & 3739 & 0.2\% \\
OpenSubtitles & 632485 & 5754 & 0.9\% \\
OpenWebText2 & 32333654 & 136914 & 0.4\% \\
PhilPapers & 63875 & 2324 & 0.4\% \\
Pile-CC & 52441354 & 19928 &  \\
PubMed Abstracts & 29329202 & 2312 &  \\
PubMed Central & 5679903 & 4230 & 0.1\% \\
StackExchange & 29529008 & 2072 &  \\
USPTO Backgrounds & 11123325 & 80088 & 0.7\% \\
Ubuntu IRC & 20067 & 10 &  \\
Wikipedia (en) & 16939503 & 45052 & 0.3\% \\
YoutubeSubtitles & 328030 & 871 & 0.3\% \\
\midrule
\textbf{Total} & \textbf{210607728} & \textbf{1296376} & \textbf{0.6\%} \\
\bottomrule
\end{tabular}
}%
\end{minipage}
\caption{
    Statistics of de-contamination in RedPajama (left) and Pile's training set (right).
}
\label{tab:bff_stats}
\end{table*}

%% file: floats/analysis_human_ngram_more.tex
\begin{figure*}[!h]
\centering
\includegraphics[width=0.5 \textwidth]{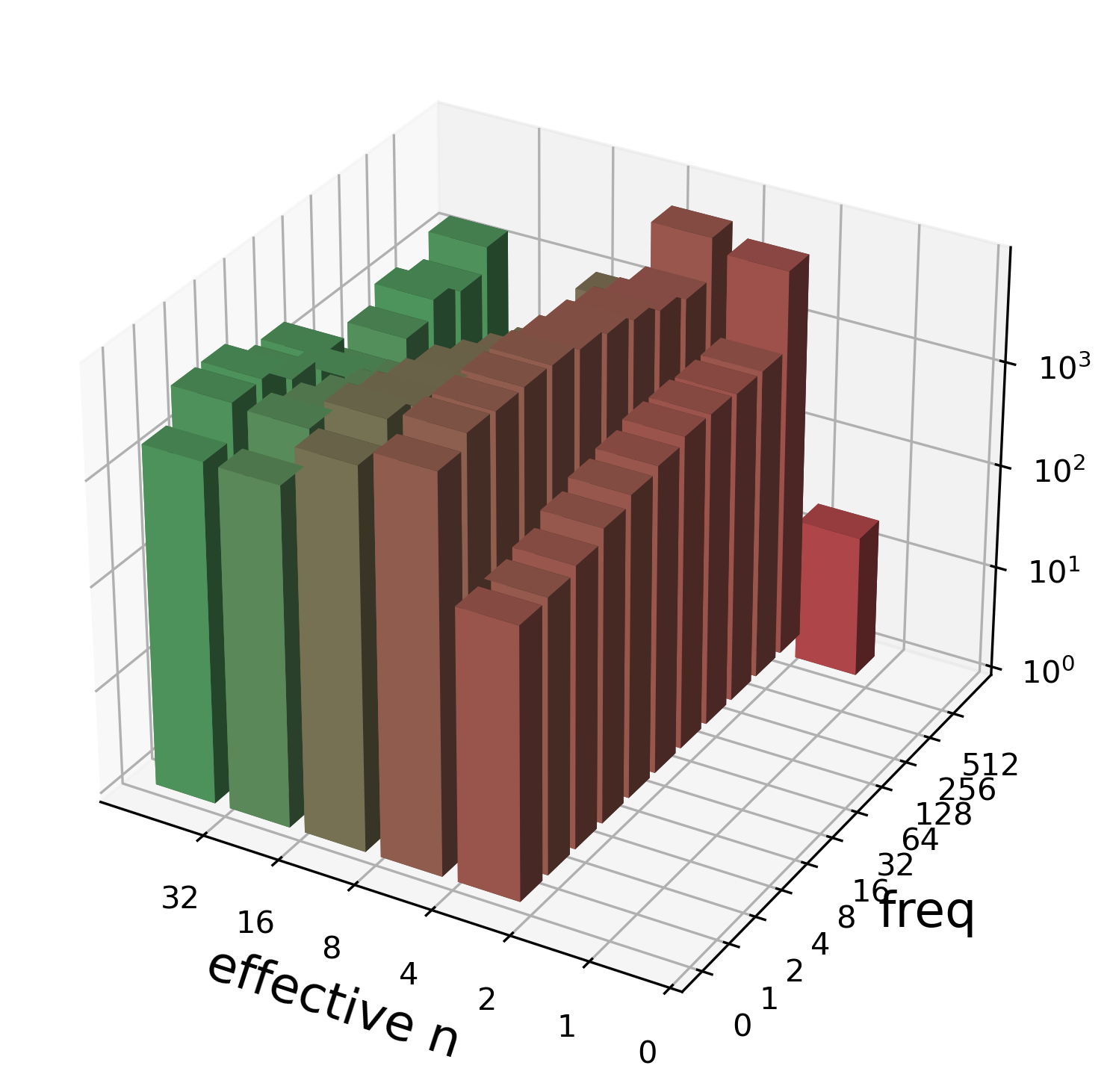}
\caption{
    Token-wise agreement between human-generated text and \methodname{}, broken down by ``effective $n$'' and frequency of the corresponding longest suffix in the reference data.
    The height of each bar represents token count, and the color represents agreement (red is 0.0, green is 1.0).
}
\label{fig:analysis_human_ngram_more}
\end{figure*}

%% file: floats/analysis_human_neural_ngram_more.tex
\begin{figure*}[!t]
\centering
\includegraphics[width=0.40 \textwidth]{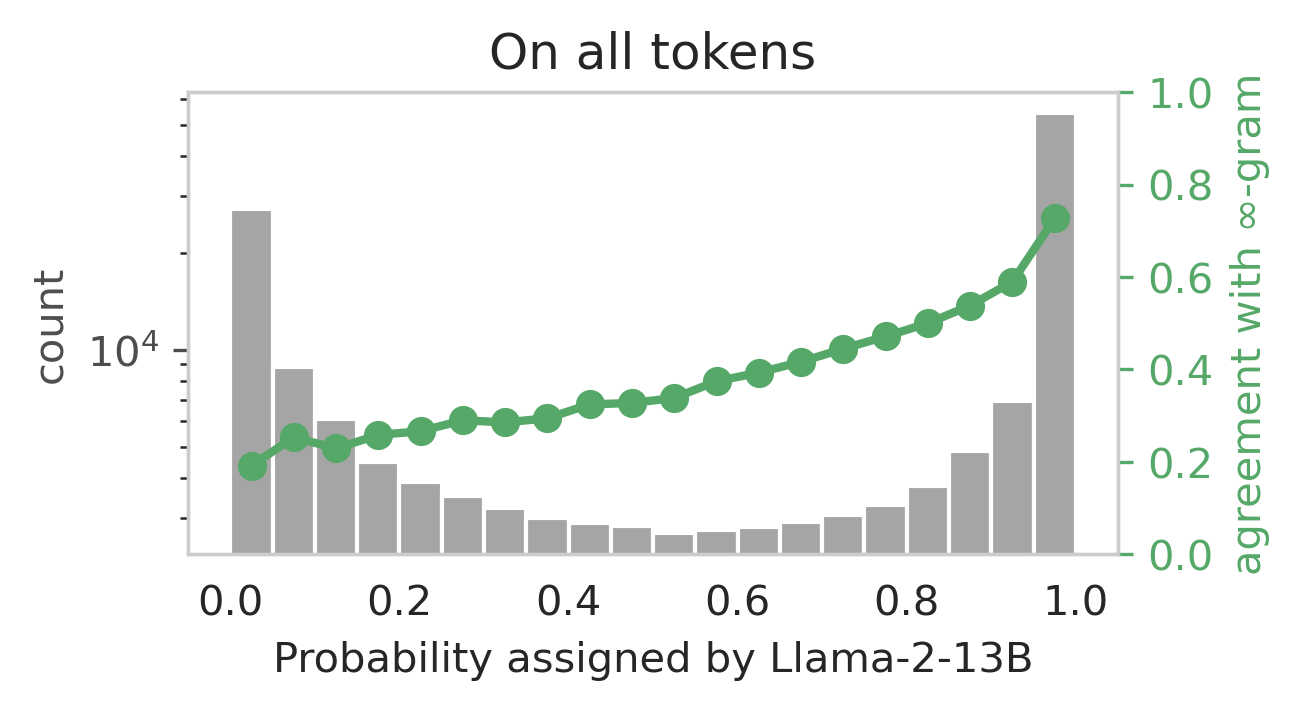}
\includegraphics[width=0.40 \textwidth]{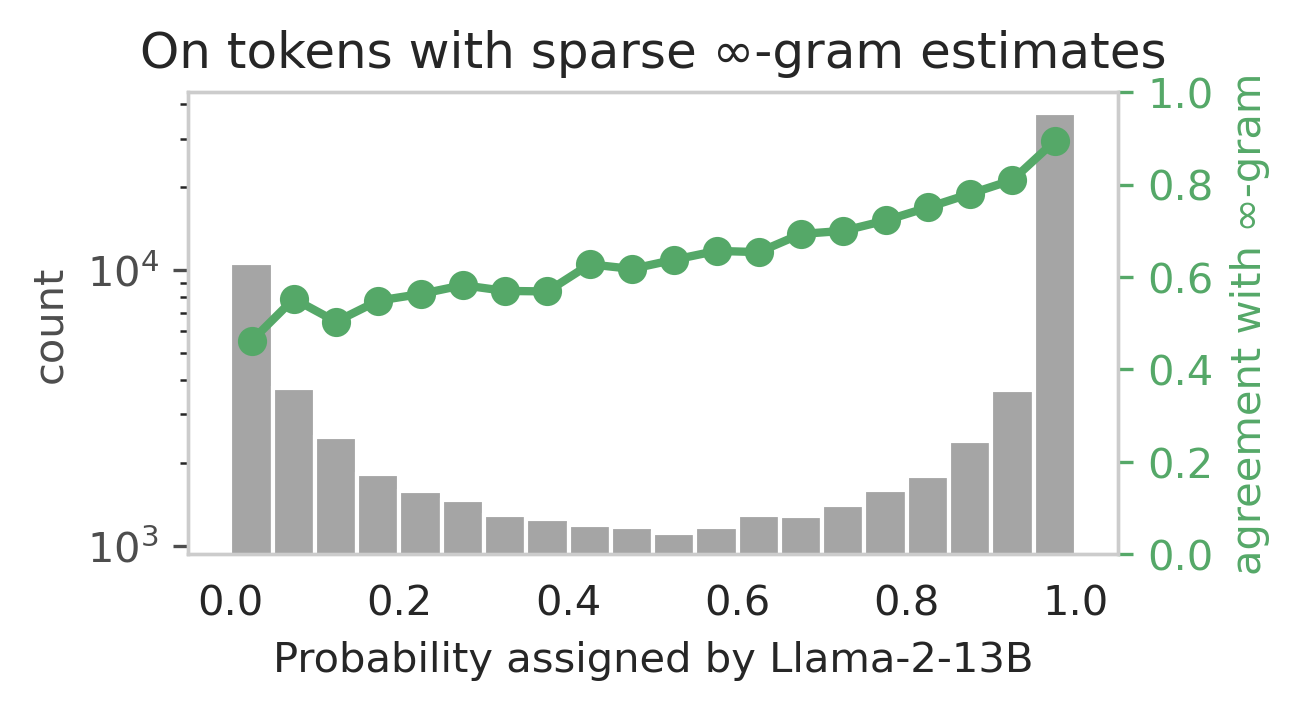}
\includegraphics[width=0.40 \textwidth]{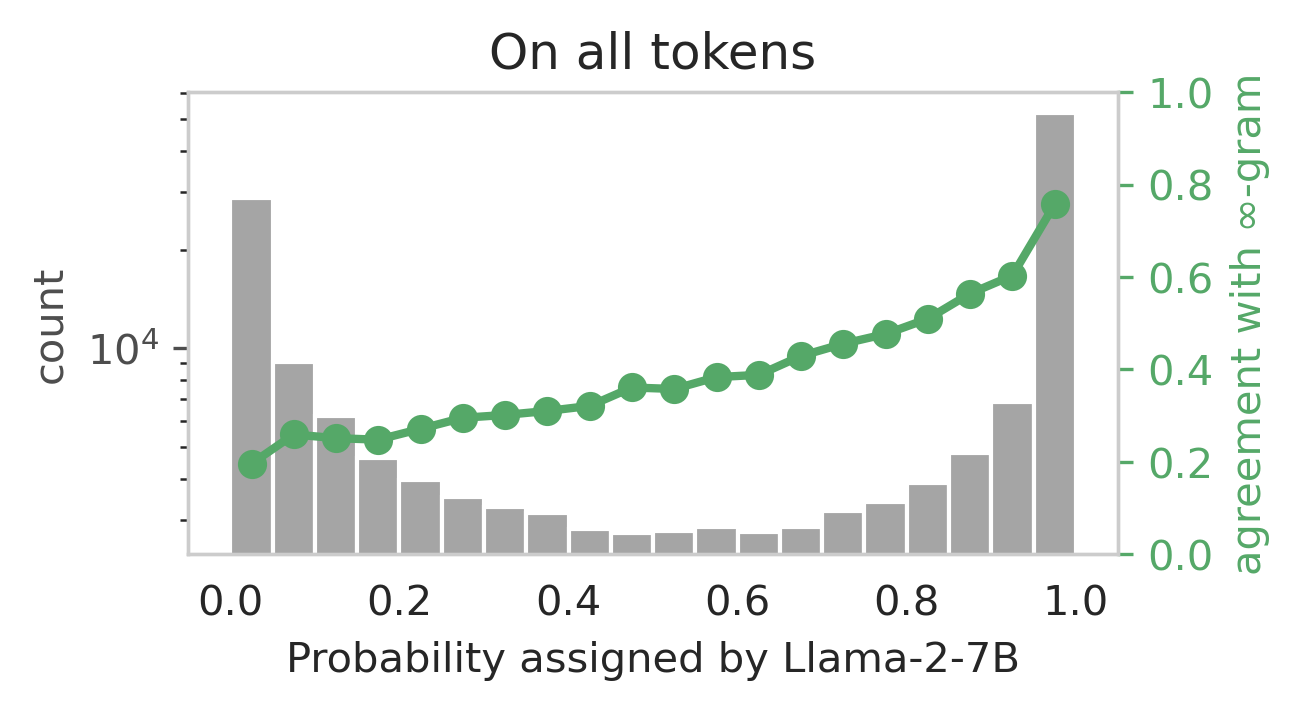}
\includegraphics[width=0.40 \textwidth]{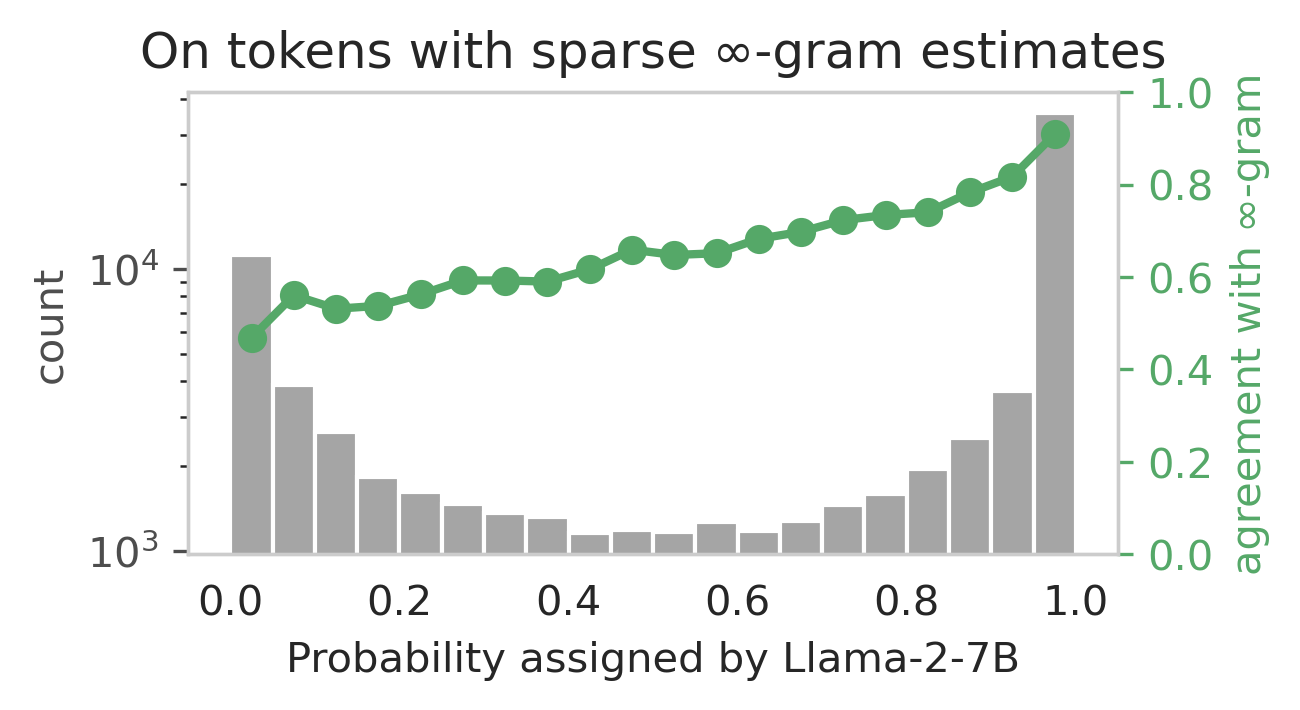}
\caption{
    Continuation of \autoref{fig:analysis_human_neural_ngram}, with results on Llama-2-13b/7b.
}
\label{fig:analysis_human_neural_ngram_more}
\end{figure*}

%% file: floats/analysis_machine_ngram_more.tex
\begin{figure*}[!t]
\includegraphics[width=0.325 \textwidth]{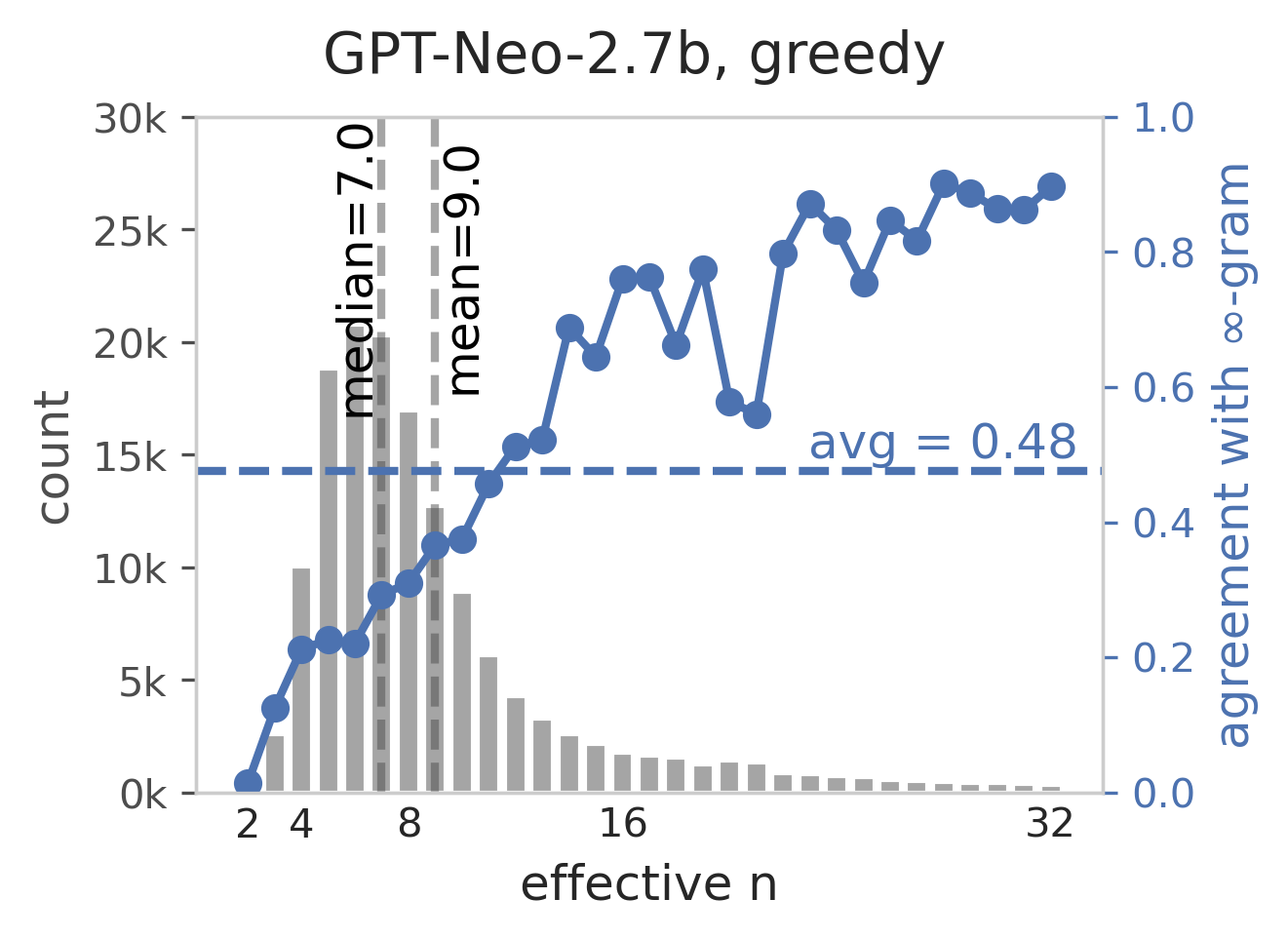}
\includegraphics[width=0.325 \textwidth]{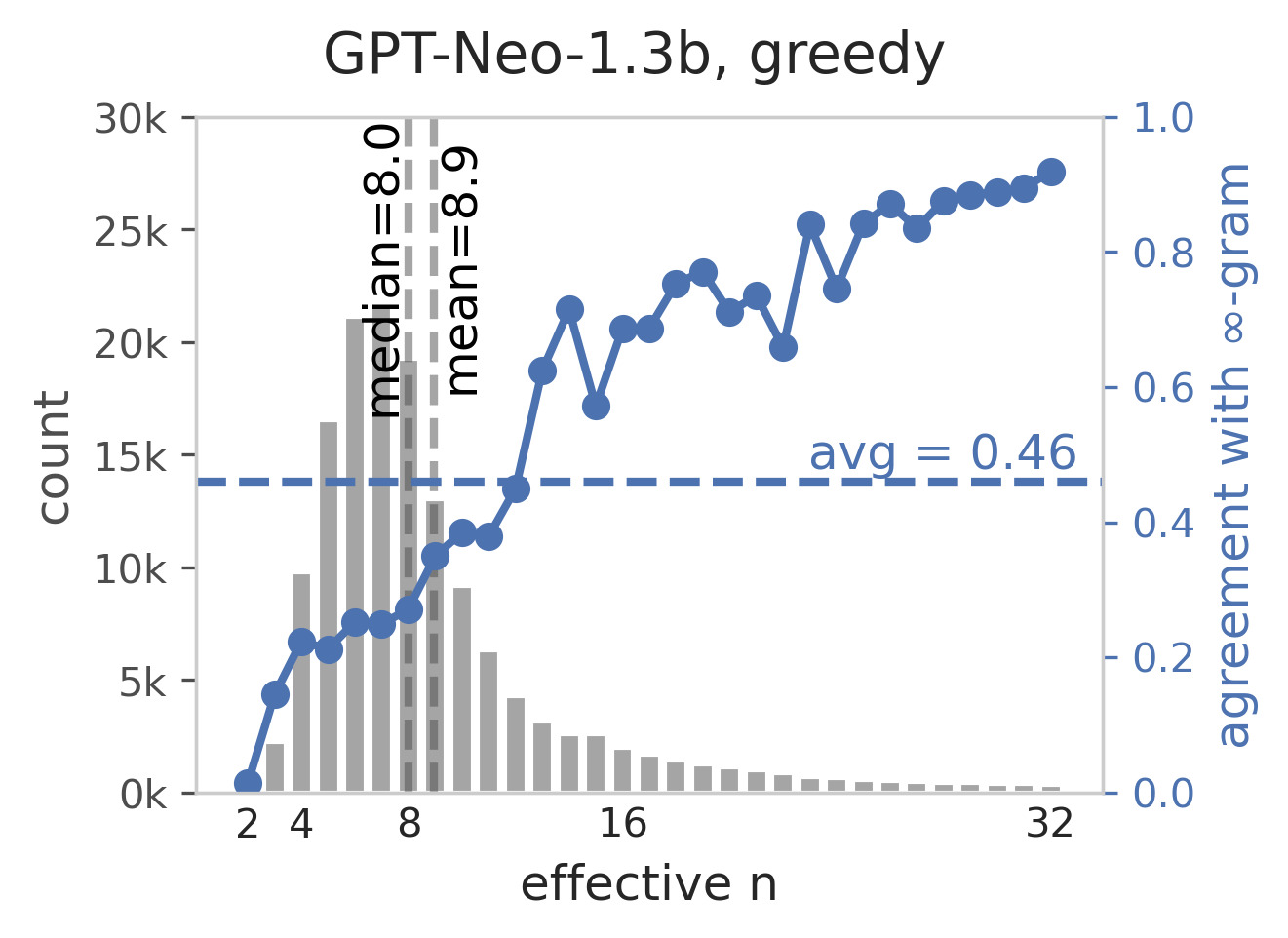}
\includegraphics[width=0.325 \textwidth]{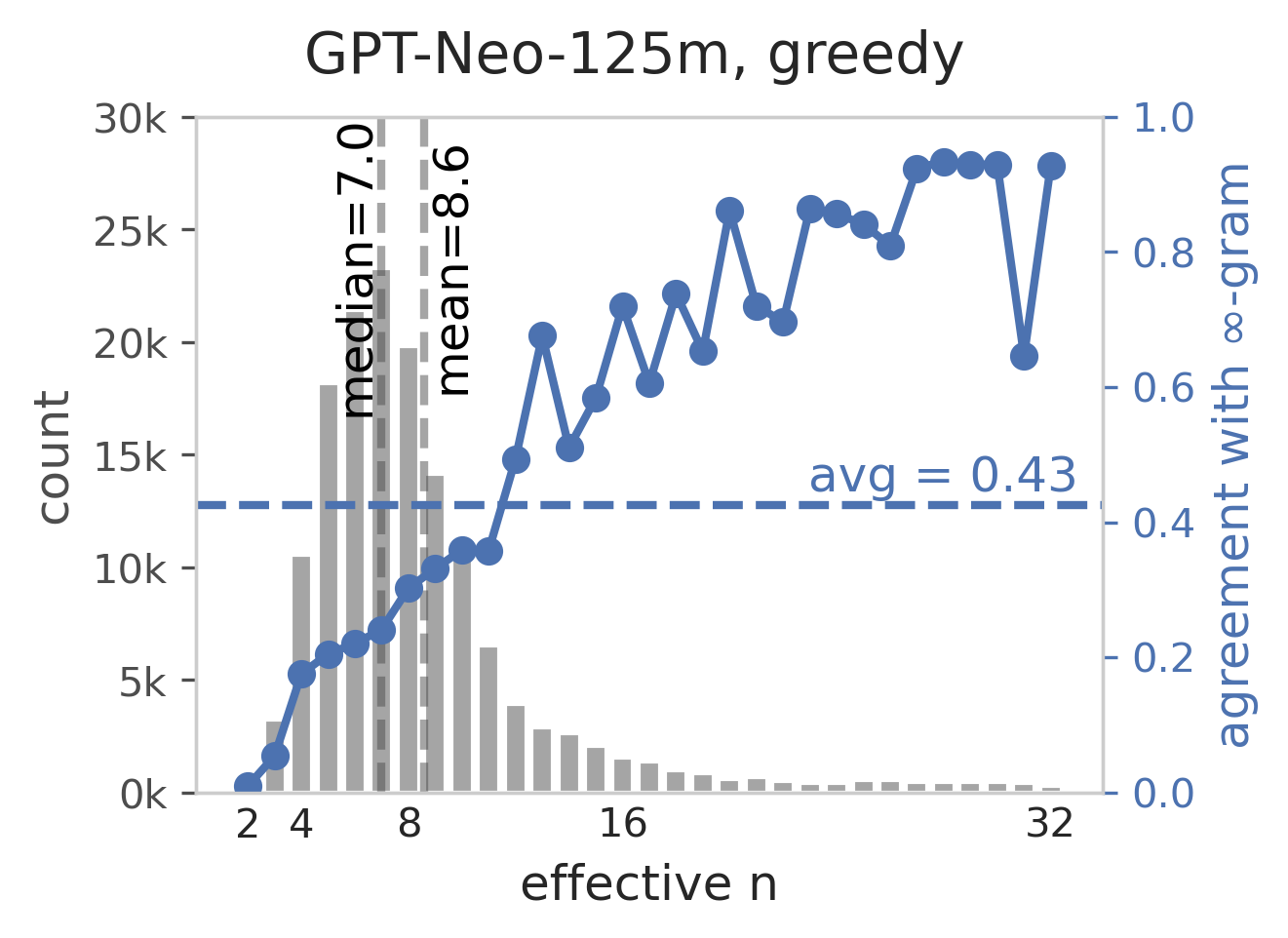}
\caption{
    Continuation of \autoref{fig:analysis_machine_ngram}, with results on GPT-Neo models.
    Token-wise agreement between machine-generated text and \methodname{}.
    All tokens are considered.
}
\label{fig:analysis_machine_ngram_more}
\end{figure*}

%% file: floats/result_timeshift.tex
\begin{table}[!t]
\small 
\centering
\setlength{\tabcolsep}{2pt}
\resizebox{0.5 \textwidth}{!}{%
\begin{tabular}{p{50pt} rrl rrl}
\toprule
\multirow{2}{*}{\parbox{40pt}{\textbf{Eval Data} (Wikipedia)}} & \multicolumn{3}{c}{simple interpolation} & \multicolumn{3}{c}{w/ Random Forest} \\
\cmidrule(lr){2-4} \cmidrule(lr){5-7}
& \textbf{Neural} & \multicolumn{2}{c}{\textbf{+ \methodname{}}} & \textbf{Neural} & \multicolumn{2}{c}{\textbf{+ \methodname{}}} \\
\midrule
April 2023  & 5.64 & \textbf{5.48} & \rel{3}  & 5.86 & \textbf{4.89} & \rel{20} \\
May 2023    & 5.43 & \textbf{5.27} & \rel{4}  & 6.01 & \textbf{5.70} & \rel{ 6} \\
June 2023   & 5.49 & \textbf{5.21} & \rel{6}  & 5.69 & \textbf{4.87} & \rel{17} \\
July 2023   & 4.93 &         4.93  & \rel{0}  & 4.91 & \textbf{4.78} & \rel{ 3} \\
August 2023 & 4.64 & \textbf{4.46} & \rel{5}  & 4.81 & \textbf{4.50} & \rel{ 8} \\
\bottomrule
\end{tabular}
}%
\caption{
    Evaluation on time-shifted data.
    The evaluation data is taken from newly-added Wikipedia articles since April 2023, which is after the creation of both the Pile and RedPajama.
    The neural model is Llama-2 (13B), and the \methodname{} reference data is Pile + RPJ.
}
\label{tab:result_timeshift}
\end{table}

%% file: floats/ablation_scale.tex
\begin{figure*}[!t]
\centering
\includegraphics[width=\textwidth]{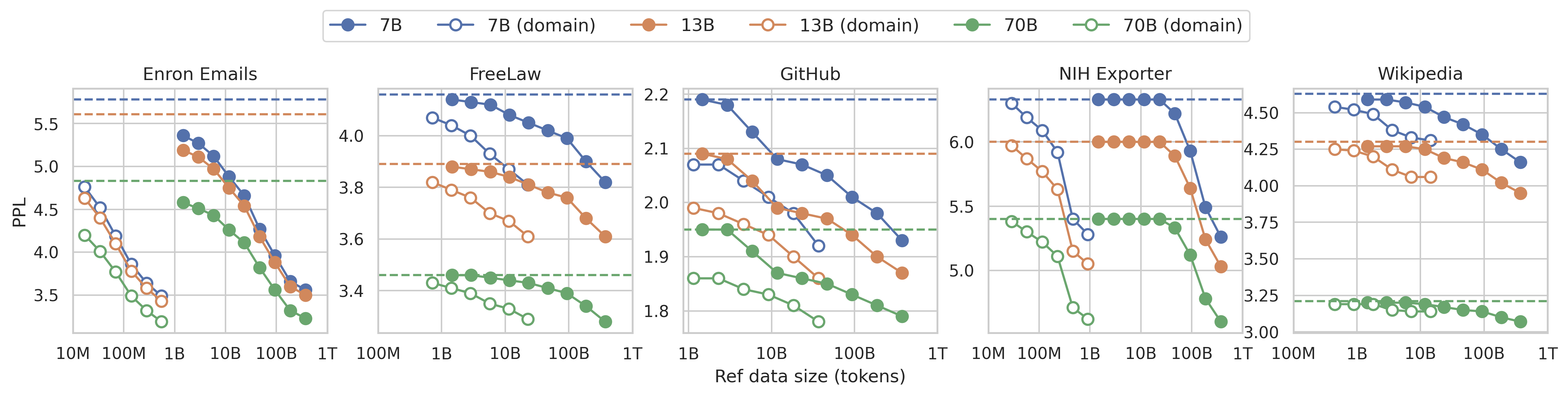}
\vspace{-8pt}
\caption{
    Impact of scaling the datastore of the \methodname{}, all using the Llama-2 models (7B, 13B, and 70B) as neural LMs, and the Pile as the reference data.
    - - -: neural LM only (baseline).
    $\bullet$: \methodname{} uses the full Pile; $\circ$: \methodname{} uses only the in-domain portion of the Pile.
    \textbf{Gains increase consistently as the datastore scales.}
}
\vspace{-12pt}
\label{fig:ablation_scale}
\end{figure*}

%% file: floats/related.tex
\begin{table*}[!t]
\centering
\setlength{\tabcolsep}{3pt}
\resizebox{\linewidth}{!}{%
\begin{tabular}{lcccc}
\toprule
\textbf{Method} & \textbf{\# tokens $(\uparrow)$} & \textbf{\# entries $(\uparrow)$} & \textbf{Storage usage $(\downarrow)$} & \textbf{max $n$} \\
\midrule
\multicolumn{4}{l}{\textbf{\em Vector-based index}} \\
\retro{} \citep{borgeaud2022improving} & 1.8 T & $2.8 \times 10^{10}$ & 432 TB (16$k$ bytes / entry) & -- \\
Atlas \citep{izacard2022few} & 27 B & $4 \times 10^8$ & 200 GB (8 bytes / entry) & -- \\
\knn{}-LM \citep{khandelwal2020generalization} & 3 B & $3 \times 10^9$ & 200 GB (64 bytes / entry) & -- \\
NPM \citep{min2022nonparametric} & 1 B & $1 \times 10^9$ & 1.4 TB ($\sim 2 k$ bytes / entry) & -- \\
\midrule
\multicolumn{4}{l}{\textbf{\em $n$-gram-based index}} \\
\citet{Brants2007LargeLM} & 2 T & $3 \times 10^{11}$ & unreported & 5 \\
Google's \citep{franz2006all} & 1 T & $3.8 \times 10^9$ & 24 GB & 5 \\
Google Books Ngram \citep{Aiden2011QuantitativeAO} & 500 B & unreported & unreported & 5 \\
\citet{Stehouwer2010UsingSA} & 90 M & unreported & unreported & $\infty$ \\
\citet{Kennington2012SuffixTA} & 3 M & $5 \times 10^{12}$ & 330 MB (110 bytes / token) & $\infty$ \\
\citet{Shareghi2015CompactEA} & 9 B & $8 \times 10^{18}$ & 63 GB (7 bytes / token) & $\infty$ \\
\midrule
\infinigram{} (ours) & 5 T & $1 \times 10^{25}$ & 35 TB (7 bytes / token) & $\infty$ \\
\bottomrule
\end{tabular}
}%
\caption{
    Comparison with other nonparametric language modeling methods.
    \textbf{\# tokens}: number of tokens in the inference-time reference data.
    \textbf{\# entries}: number of representations (counts) in the index.
    \textbf{max $n$}: maximum number of context tokens considered.
    For \infinigram{}, we consider the combination of Pile-train and RedPajama as reference data.
}
\label{tab:related}
\end{table*}

%% file: floats/queries/1.tex
\begin{figure}[!h]
\centering
\includegraphics[width=\textwidth]{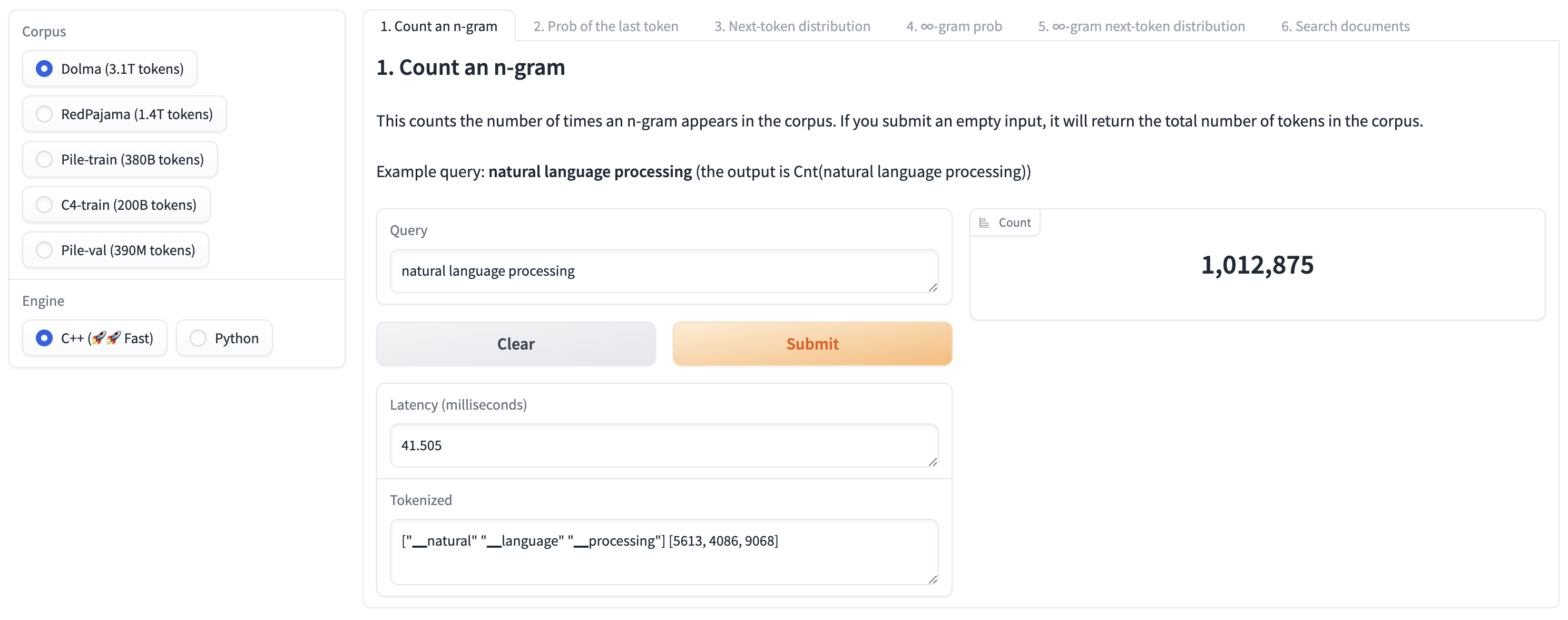}
\caption{Example for query type 1 (\textsc{Count}): Counting an \ngram{}.}
\label{fig:queries_1}
\end{figure}

%% file: floats/queries/2.tex
\begin{figure}[!h]
\centering
\includegraphics[width=\textwidth]{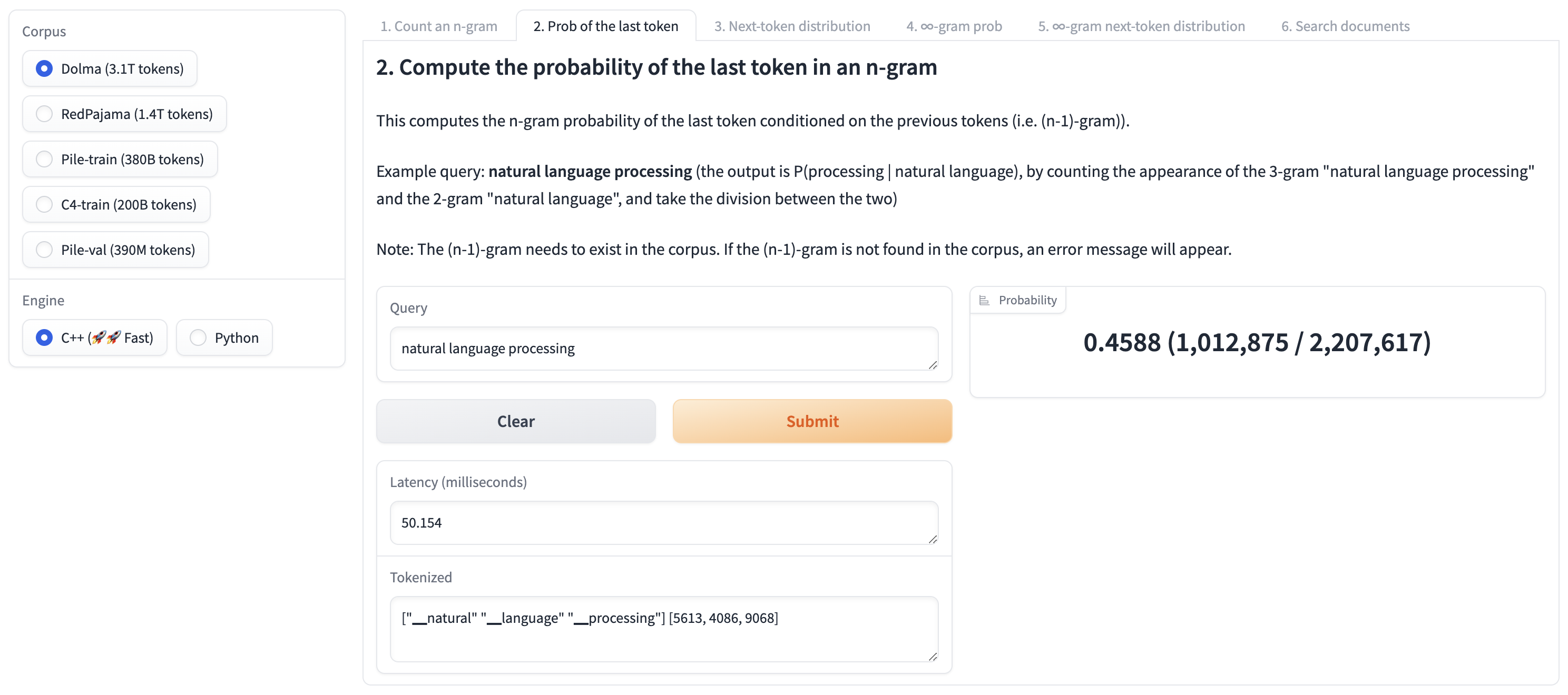}
\caption{Example for query type 2 (\textsc{NgramProb}): Computing a token probability from \ngram{} LM (with given $n$, no backoff).}
\label{fig:queries_2}
\end{figure}

%% file: floats/queries/3.tex
\begin{figure}[!h]
\centering
\includegraphics[width=\textwidth]{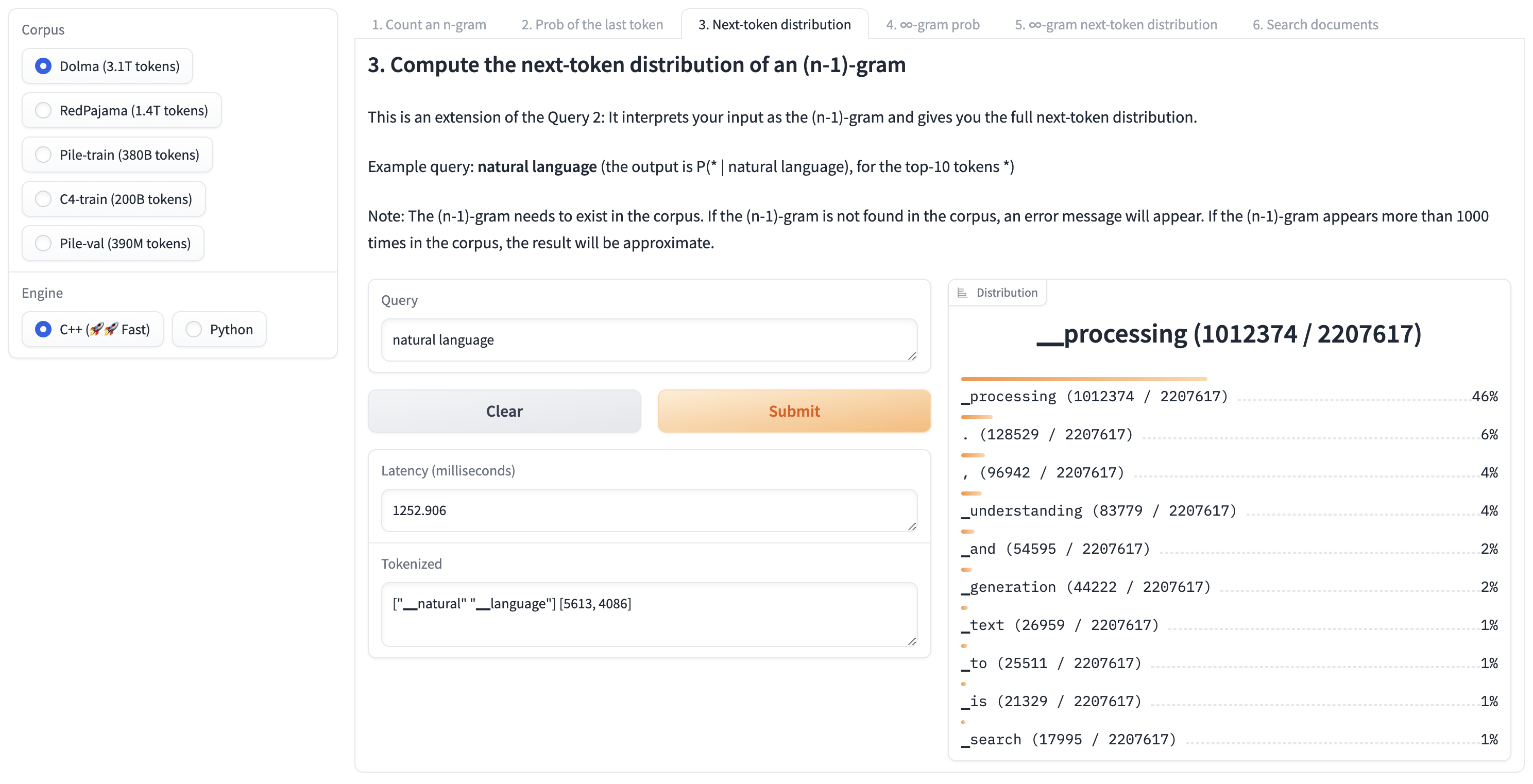}
\caption{Example for query type 3 (\textsc{NgramDist}): Computing the full next-token distribution from \ngram{} LM. Due to space limits, only top-10 tokens are shown.}
\label{fig:queries_3}
\end{figure}

%% file: floats/queries/4.tex
\begin{figure}[!h]
\centering
\includegraphics[width=\textwidth]{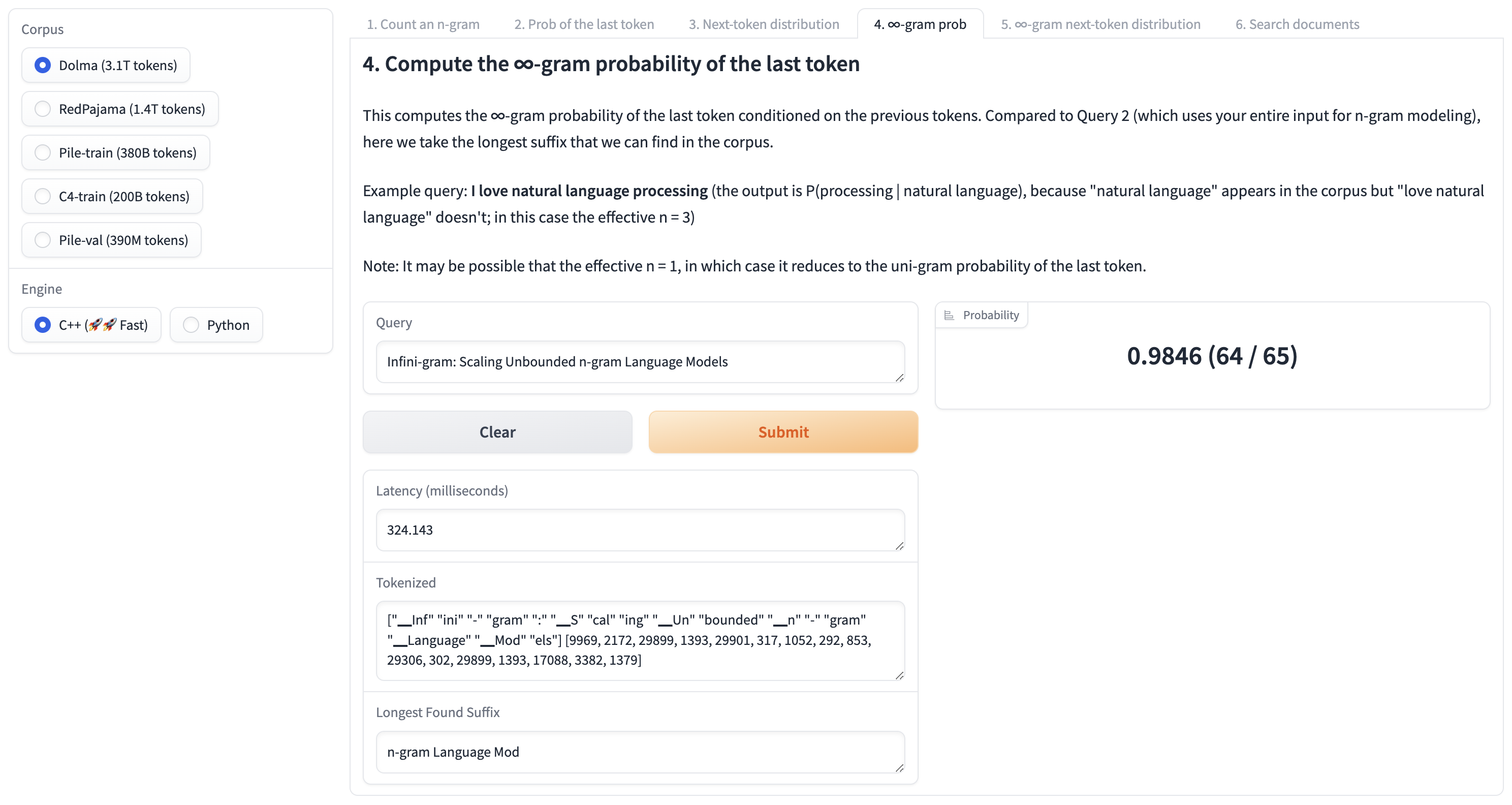}
\caption{Example for query type 4 (\textsc{InfgramProb}): Computing a token probability from \methodname{} LM.}
\label{fig:queries_4}
\end{figure}

%% file: floats/queries/5.tex
\begin{figure}[!h]
\centering
\includegraphics[width=\textwidth]{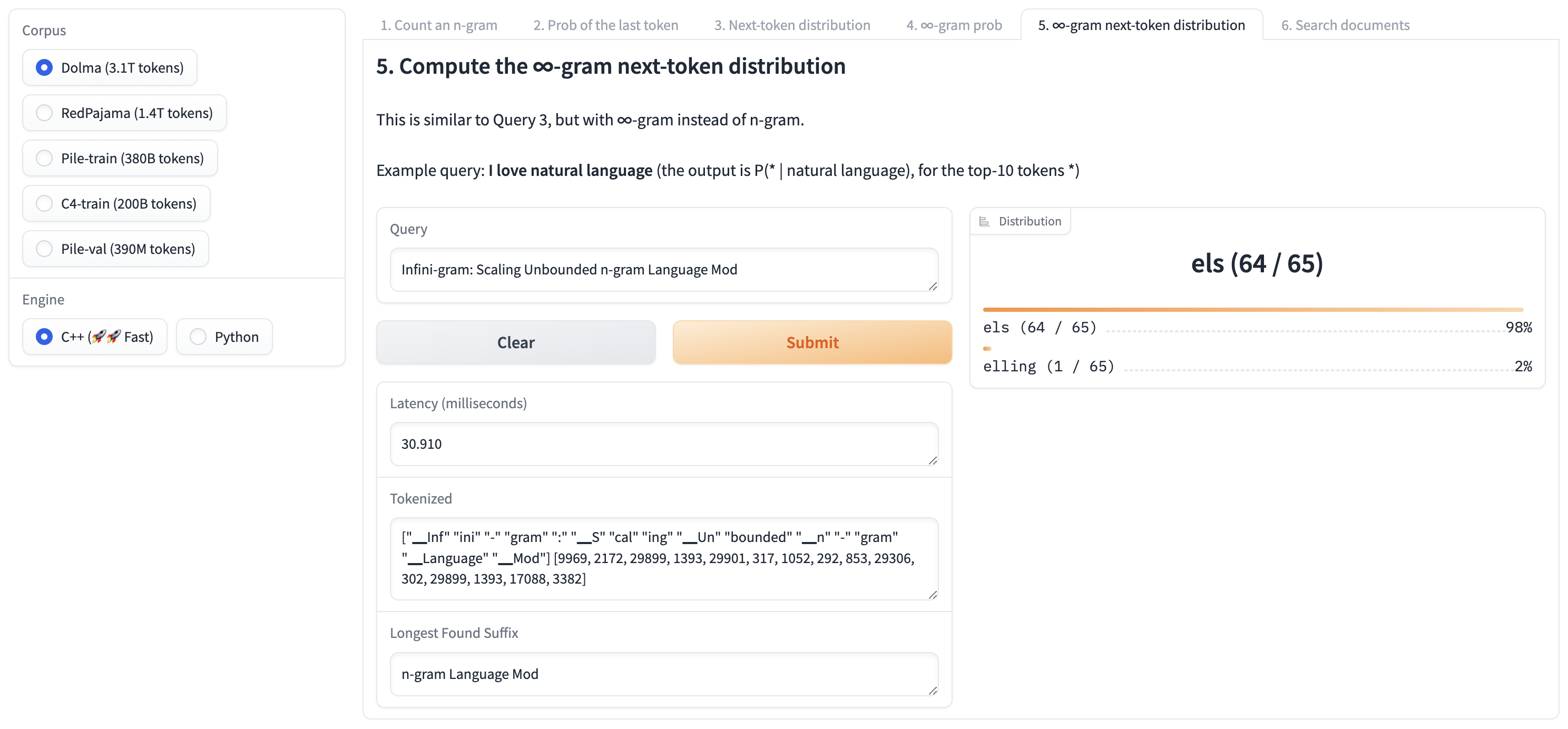}
\caption{Example for query type 5 (\textsc{InfgramDist}): Computing the full next-token distribution from \methodname{} LM. Due to space limits, only top-10 tokens are shown.}
\label{fig:queries_5}
\end{figure}

%% file: floats/queries/6.tex
\begin{figure}[!h]
\centering
\includegraphics[width=\textwidth]{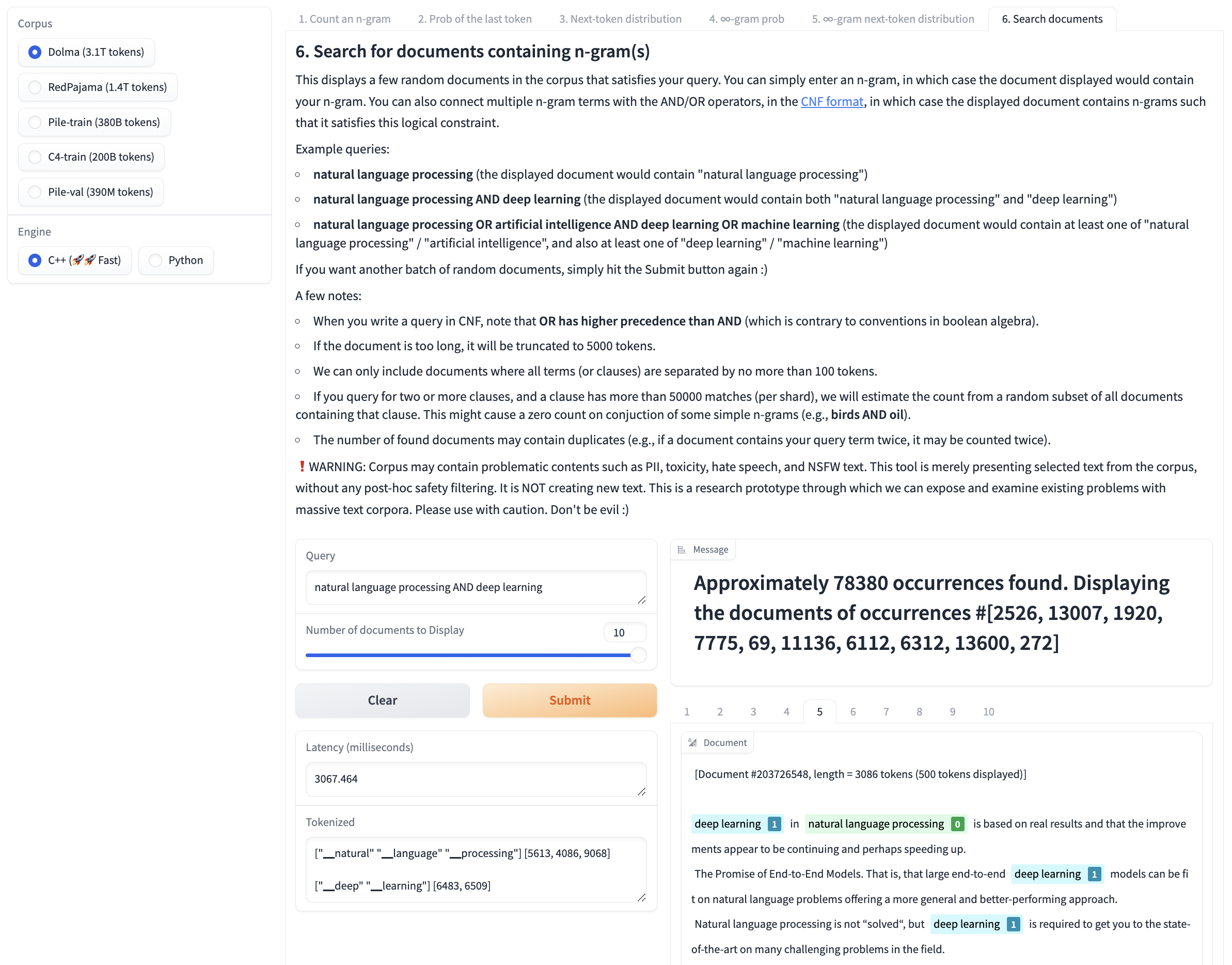}
\caption{Example for query type 6 (\textsc{SearchDoc}): Returning documents containing an \ngram{}, or a CNF logical expression of \ngram{} terms, connected with AND's and/or OR's.}
\label{fig:queries_6}
\end{figure}